\documentclass[letterpaper]{article} % DO NOT CHANGE THIS
\usepackage{aaai2026}  % DO NOT CHANGE THIS
\usepackage{times}  % DO NOT CHANGE THIS
\usepackage{helvet}  % DO NOT CHANGE THIS
\usepackage{courier}  % DO NOT CHANGE THIS
\usepackage[hyphens]{url}  % DO NOT CHANGE THIS
\usepackage{graphicx} % DO NOT CHANGE THIS
\usepackage{natbib}  % DO NOT CHANGE THIS AND DO NOT ADD ANY OPTIONS TO IT
\usepackage{caption} % DO NOT CHANGE THIS AND DO NOT ADD ANY OPTIONS TO IT
\usepackage{algorithm}
\usepackage{newfloat}
\usepackage{listings}
\usepackage{microtype}
\usepackage{booktabs} % for professional tables

\usepackage{mathtools}
\usepackage{amsthm}
\usepackage{amsmath,amsfonts,bm}

\def\eqref#1{equation~\ref{#1}}
\def\1{\bm{1}}

\DeclareMathAlphabet{\mathsfit}{\encodingdefault}{\sfdefault}{m}{sl}
\SetMathAlphabet{\mathsfit}{bold}{\encodingdefault}{\sfdefault}{bx}{n}

\usepackage{graphicx}
\usepackage{subfig}
\usepackage{url}
\usepackage{algorithm}
\usepackage{algcompatible}
\usepackage{algpseudocode}

\usepackage{amssymb}
\usepackage{amsmath}
\usepackage{multirow}
\usepackage{pifont}

\newtheorem{innercustomthm}{Theorem}

\newtheorem{innercustomlem}{Lemma}
\newenvironment{customlem}[1]
  {\renewcommand\theinnercustomthm{#1.}\innercustomlem}
  {\endinnercustomthm}

\usepackage{newfloat}
\usepackage{listings}
\usepackage[dvipsnames]{xcolor}
\usepackage{xcolor}

\DeclareCaptionStyle{ruled}{labelfont=normalfont,labelsep=colon,strut=off} % DO NOT CHANGE THIS
\floatstyle{ruled}
\newfloat{listing}{tb}{lst}{}
\floatname{listing}{Listing}
\title{Eliciting Chain-of-Thought in Base LLMs via Gradient-Based \\ Representation Optimization}

\author {
    Zijian Wang,
    Yanxiang Ma,
    Chang Xu
}
\affiliations {
    School of Computer Science, The University of Sydney\\
    zwan0998@uni.sydney.edu.au, yama9404@uni.sydney.edu.au, c.xu@sydney.edu.au
}

\usepackage{bibentry}
\begin{document}

\maketitle

\begin{abstract}
% Abstract section
% This file contains the abstract content for the AAAI paper

% Chain of Thought (CoT) reasoning is a critical capability for Large Language Models (LLMs) in solving complex tasks. 
% Recent research indicates that reasoning abilities are encoded within LLMs' latent space and can be elicited through activation editing techniques, such as linear vector steering. 
% However, such vector arithmetic approaches are rigid, lacking principled objectives to guide the editing process, potentially causing distribution shift and unintended side effects associated with unconstrained steering scale.
% In this paper, we propose a gradient-based activation optimization method to elicit CoT reasoning from a principled probabilistic generation perspective. 
% Deriving an analytical objective from Bayesian learning principles, our approach balances a likelihood term that directs hidden states toward reasoning capabilities with a prior term that preserves proximity to the original distribution.
% Experimental results across multiple reasoning benchmarks and LLMs demonstrate that our method consistently outperforms previous vector arithmetic approaches while ensuring activations remain within the model's natural distribution, avoiding the negative effects caused by potential distribution shift.

Chain-of-Thought (CoT) reasoning is a critical capability for large language models (LLMs), enabling them to tackle complex multi-step tasks. 
While base LLMs, pre-trained on general text corpora, often struggle with reasoning due to a lack of specialized training, recent studies reveal their latent reasoning potential tied to hidden states. 
However, existing hidden state manipulation methods, such as linear activation steering, suffer from limitations due to their rigid and unconstrained nature, often leading to distribution shifts and degraded text quality. 
In this work, we propose a novel approach for eliciting CoT reasoning from base LLMs through hidden state manipulation grounded in probabilistic conditional generation. 
By reformulating the challenge as an optimization problem with a balanced likelihood and prior regularization framework, our method guides hidden states toward reasoning-oriented trajectories while preserving linguistic coherence. 
Extensive evaluations across mathematical, commonsense, and logical reasoning benchmarks demonstrate that our approach consistently outperforms existing steering methods, offering a theoretically principled and effective solution for enhancing reasoning capabilities in base LLMs.
\end{abstract}

\section{Introduction}

Chain-of-Thought (CoT) reasoning is a vital capability for large language models (LLMs), enabling them to break down complex problems into intermediate steps for improved performance on complex tasks \cite{wei2022chain, kojima2022large}. 
Models pre-trained solely on general text corpora, known as \textit{Base LLMs}, often excel in fluency language generation but struggle with complex reasoning due to a lack of specialized training. 
In contrast, advanced models, termed \textit{Reasoning LLMs}, incorporate reasoning-focused data (e.g., STEM, coding, synthetic examples) during pre-training and undergo extensive post-training, such as instruction fine-tuning and reinforcement learning, to enhance reasoning abilities \cite{guo2025deepseek, yang2025qwen3}.

Intriguingly, recent studies have uncovered latent reasoning potential within purely pre-trained Base LLMs.
For instance,~\cite{wang2024chain} demonstrates that, when sampling multiple responses from an LLM for a given problem, CoT-style answers are frequently present among the diverse outputs, even though the default greedy decoding yields only direct answers.
Complementarily,~\cite{wang2025thoughtprobe, bi2025cot} shows that the presence of CoT can be effectively assessed from the model's hidden states, revealing a strong correlation between intrinsic reasoning capabilities and these internal representations. 
These findings underscore that pre-trained LLMs contain unlocked reasoning abilities tied to their hidden states, prompting efforts to enhance such models by directly manipulating these representations. 
Existing approaches primarily rely on linear activation steering, which computes and applies a pre-defined control vector with fixed steering strength in a single step to redirect the model's generation toward reasoning-oriented trajectories~\cite{hojer2025improving, tang2025unlocking, hong2025reasoning}. 

Despite its intuitive appeal and simplicity, linear activation steering suffers from significant limitations due to its inflexible framework. 
It fails to adapt to instance-specific variations and lacks a clear optimization objective to direct the steering process. 
Furthermore, it applies unbounded steering strength without well-defined constraints. 
As a result, this approach can disrupt the latent manifold structure, causing distribution shifts and degrading text quality, which ultimately leads to deviations from the model's natural language generation capabilities \cite{da2025steering, von2024language, li2023inference}.
Therefore, we pose the question: \textit{Can we solely manipulate the hidden states to elicit CoT reasoning from a base LLM, without distorting its language quality?}

\begin{figure*}[t]
    \begin{center}
    \centerline{\includegraphics[scale = 0.5]{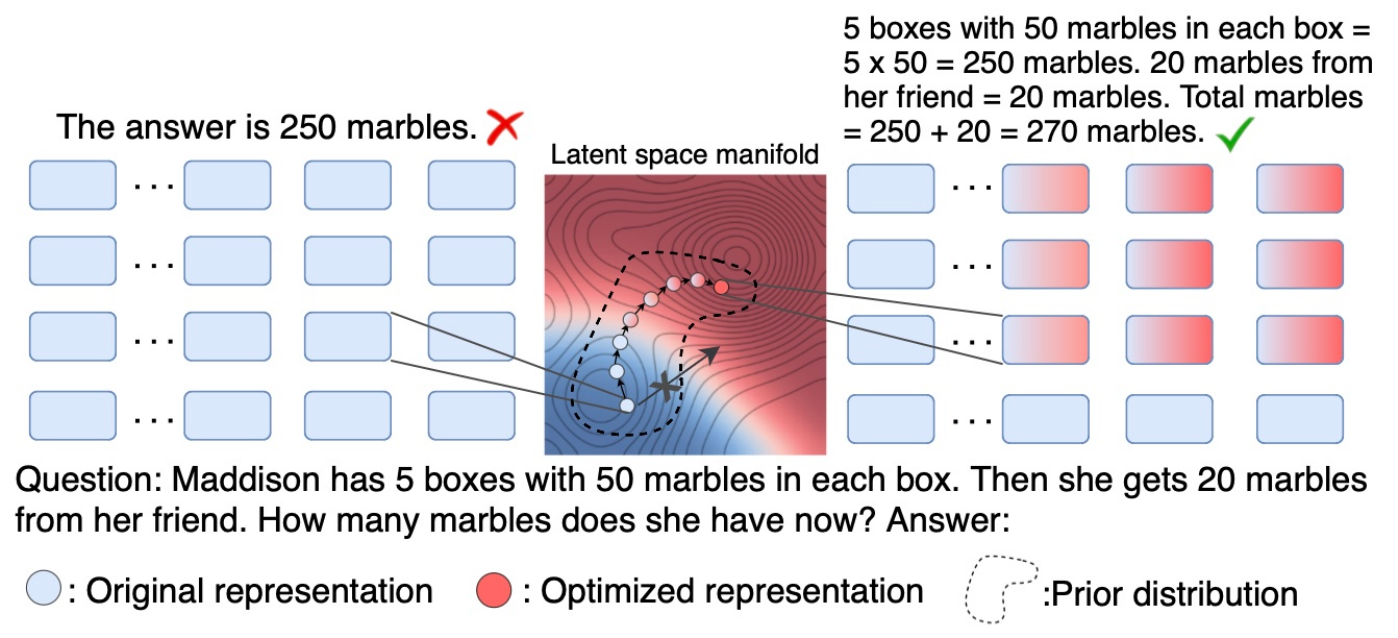}}
    \caption{Our method optimizes hidden states within the base LLM's latent space to elicit Chain-of-Thought reasoning. Before optimization, the model produces direct answers without reasoning steps; after optimization, it generates chain of thought content.
    Unlike linear activation steering that may disrupt the language prior without regulations, our approach implements guided optimization with principled constraints.}
    \vspace{-5mm}
    \label{Intro}
    \end{center}
  \end{figure*}

In this paper, we propose an alternative hidden states manipulation approach to elicit CoT reasoning from base LLMs based on probabilistic conditional generation principles.
The foundation of our method lies in establishing a connection between hidden states manipulation and latent variable learning within a conditional text generation framework, which requires sampling appropriate hidden states from an intractable posterior distribution that satisfy the desired condition.

Inspired by Bayesian posterior estimation \cite{welling2011bayesian}, we reformulate this sampling challenge as an optimization problem comprising two essential components: a likelihood term and a prior term.
The likelihood term, implemented as a pre-trained classifier, quantifies the probability that hidden states satisfy the target condition, thereby guiding these states toward the desired reasoning mode.
Concurrently, the prior term functions as a regularization mechanism, constraining hidden states to remain proximate to the original manifold structure, thus preserving linguistic coherence in the generated text.
These complementary objectives are carefully balanced to ensure optimization convergence while maintaining both reasoning effectiveness and linguistic fluency.
During LLM inference, our method employs gradient-based optimization to compute hidden states that maximize this composite objective function.
This approach effectively navigates the representation space to satisfy the CoT condition while the prior regularization avoids the distribution shift and degradation of text quality.

We evaluate our method on several reasoning benchmarks, including mathematical, commonsense, and logical reasoning tasks. 
Our method is proven to be effective in eliciting CoT reasoning capabilities in base LLMs compared to vector arithmetic methods on multiple major benchmarks.

The main contributions of this work are:

1: We propose a new hidden states manipulation method based on probabilistic conditional generation theory to elicit CoT reasoning from base LLMs. 
Our approach is theoretically grounded and principled, offering advantages over conventional  rigid vector arithmetic methods.

2: We develop a comprehensive optimization framework that balances likelihood maximization with prior regularization, effectively guiding hidden states toward reasoning capabilities while preserving representational integrity. We establish theoretical bounds to guarantee this trade-off.

3:  We demonstrate our method's effectiveness through extensive evaluation across diverse reasoning benchmarks, consistently outperforming existing hidden states steering approaches.

\section{Background}

\subsection{LLM Text Generation}
LLMs generate text in an autoregressive manner. Given a sequence of tokens $X = (x_1, x_2, \cdots, x_n)$ with length $n$, the probability of the sequence $p(X)$ is factorized as:
\begin{equation}
p(X) = \prod_{i=1}^{n} p(x_i|x_0, \cdots, x_{i-1}).
\label{eq1:llm_generation}
\end{equation}
In transformer architectures, the contextual information from previous tokens is encoded into hidden states through multiple layers of self-attention and feed-forward networks. 
For a token at position $i$, the hidden state $h_i$ captures the representation of the sequence processed so far $h_i = f_{\text{transformer}}(x_{<i})$.
where $f_{\text{transformer}}$ represents the transformation through the model's layers, and $x_{<i} = (x_0, x_1, \ldots, x_{i-1})$ is the input context. 
We omit the layer index of $h_i$ for simplicity.

The distribution for each token $p(x_i|x_0, \cdots, x_{i-1})$ can then be expressed by the hidden states as
\begin{equation}
p(x_i|x_0, \cdots, x_{i-1}) = p(x_i|h_i) = \text{softmax}(h_i W_o),
\label{eq2:llm_generation}
\end{equation}
where $W_o$ is the weight matrix of the output head. This formulation highlights that next-token prediction is fundamentally determined by the hidden representation $h_i$, which encodes necessary contextual information for generation.

\subsection{Conditional Text Generation}
Conditional generation aims to produce text $X$ that satisfies a specific condition $c$, formally expressed as sampling from the distribution $p(X|c)$, which is a wide theoretical paradigm in language generation, such as attribute transfer \cite{wang2019controllable} and controllable text generation \cite{dathathri2019plug}, where the condition $c$ is a high level semantic attribute.
Due to the nature of LLM forward computation, the text and hidden states are jointly defined in a distribution.
Formally, we can formulate the conditional generation process as
\begin{equation}
p(X, h|c)= p(X|h,c)p(h|c) =  p(X|h)p(h|c).
\label{eq3:conditional_generation}
\end{equation}

The equality $p(X|h,c) = p(X|h)$ holds because once the hidden state $h$ is fixed, the LLM's decoding process is fully determined and independent of the condition $c$. 
This formulation reveals that $h$ functions as a mediator between the condition $c$ and the generated text $X$.

% To generate text satisfying condition $c$, we can employ ancestral sampling along the causal path $c \rightarrow h \rightarrow X$. 
% This process involves first sampling a hidden state from $p(h|c)$ that satisfies the desired condition, followed by sampling text from $p(X|h)$ through the standard LLM decoding mechanism. 
% While the latter distribution is already parameterized by LLM architecture, the principal challenge resides in efficiently sampling from $p(h|c)$, which is generally intractable due to the high-dimensional nature of hidden states and the complex relationship between conditions and representations.

To generate text satisfying condition $c$, we can employ ancestral sampling along the causal path $c \rightarrow h \rightarrow X$. This involves two steps, first sampling a hidden state from $p(h|c)$ that meets the desired condition, then generating text from $p(X|h)$ via the decoding process of LLMs. While the LLMs architecture already parameterized the second distribution $p(X|h)$, the key challenge lies in efficiently sampling from $p(h|c)$. The sampling process is normally intractable due to the high dimensionality of hidden states and the complex condition-representation relationship.

\section{Methodology}

In this section, we present our approach to elicit chain-of-thought reasoning in base LLMs through hidden states optimization.
We first formulate the problem within a Bayesian posterior inference framework, then derive an analytical optimization objective form with two  complementary targets, and develop a gradient-based optimization method for its practical implementation.
\subsection{Problem Formulation}
In the conditional generation framework the hidden state $h$ serves as the causal mediator between CoT condition $c$ and generated text $X$. By building this framework,
we confront the fundamental challenge of sampling from the posterior distribution $p(h|c)$ to obtain hidden states that satisfy our desired condition. 
Rather than attempting sampling from the posterior distribution, we employ Maximum Posterior (MAP) estimation~\cite{welling2011bayesian}
to identify the mode of the posterior distribution, formally,
\begin{equation}
h^* = \text{argmax}_h[\log p(h|c)].
\label{eq4:map_estimation}
\end{equation}

Practically, we apply Bayesian rule to decompose the posterior distribution as $p(h|c) = \frac{p(c|h)p(h)}{p(c)}\propto p(c|h)p(h)$ and the optimization target in Eq.~\ref{eq4:map_estimation}is equivalent to:
\begin{equation}
h^* = \text{argmax}_h[\log p(c|h) + \log p(h)]
\label{eq5:optimization_target}
\end{equation}
where $p(c|h)$ is the likelihood and $p(h)$ is the prior term. 
To solve this optimization in practice, we use gradient-based methods starting from $h_0$, the original hidden state obtained from the LLM's initial forward pass on the input context.

\subsection{Deriving the Optimization Objective}
We first introduce how we model the prior and likelihood term to derive a analytical solution for the optimization target. 
And then, we revise the optimization process with an adaptive term and a regularization noise. Finally, we demonstrate the entire process of our method.

% For the prior distribution $p(h)$, we employ a Locally Gaussian Approximation \cite{kingma2013auto}, modeling it as a Gaussian centered at the original hidden state $h_0$. 
% This is expressed as $p(h) \propto \exp(-d(h, h_0))$, where probability density decreases exponentially with distance from $h_0$, and $d(h, h_0)$ represents the L2 distance measuring deviation from $h_0$, constraining optimization to remain within a reasonable neighborhood. 
% \begin{equation}
%   -\log p(h) \approx d(h, h_0) \propto \frac{1}{2} \|h - h_0\|^2
%   \label{eq6:prior_distribution}
% \end{equation}

\textbf{Modeling Prior} In the context of posterior estimation, the prior distribution \(p(h)\) serves to constrain the optimization process by penalizing large deviations from the original hidden state \(h_0\) \cite{fortuin2022priors}, ensuring that the resulting hidden state remains plausible and stable within a reasonable neighborhood.  
We model this prior as \(p(h) \propto \exp(-d(h, h_0))\), where the probability density decreases exponentially with the L2 distance \(d(h, h_0)\) from \(h_0\).
\begin{equation}
  -\log p(h) \approx d(h, h_0) \propto \frac{1}{2} \|h - h_0\|^2
  \label{eq6:prior_distribution}
\end{equation}

More theoretical justification for this L2 regularization term is provided in the appendix.

\textbf{Likelihood Term}
For the likelihood term $p(c|h)$, we implement a Multi-Layer Perceptron classifier $f_\theta(h)$ estimating the probability that hidden state $h$ exhibits the target condition $c$. 
Trained on a contrastive dataset of hidden states from both CoT and non-CoT responses (labeled positive and negative, respectively), the classifier is optimized using cross-entropy loss to effectively distinguish between reasoning and non-reasoning hidden states. 
More details are provided in the experiments.

With the above two components, we can rewrite the optimization target as:
\begin{equation}
h^* = \text{argmax}_h [\log f_\theta(h) - d(h, h_0)]
\label{eq7:optimization_objective}
\end{equation}

And we can use gradient ascent to update the hidden state with a step size $\alpha$:
\begin{equation}
h_{t+1} = h_t + \alpha \nabla_h [\log f_\theta(h_t) - d(h_t, h_0)]
\label{eq8:gradient_ascent}
\end{equation}

\textbf{Adaptive Step Size and Random Noise} Drawing from gradient descent principles~\cite{lr_decay}, we implement an adaptive step size $\alpha_t$ that dynamically decays as the $f_\theta(h_t)$ approaches the target value, ensuring both efficient convergence and optimization stability, defined as
$\alpha_t = \alpha_0 \cdot \frac{|\tau - f(h_t) |}{1+ f(h_0)}$
where $\tau$ is the target value of the optimization and $\alpha_0$ is the initial step size.
Further, to facilitate exploration beyond local minima and ensure authentic sampling from the distribution~\cite{welling2011bayesian}, we incorporate Gaussian noise $z \sim \mathcal{N}(0, I)$ into the optimization process.

Using gradient ascent with the adaptive step length and random noise, the iterative updating on $h$ at timestep $t$ can finally be formulated as
\begin{equation}
  h_{t+1} = h_t + \alpha_t \nabla_h [\log f_\theta(h_t) - \lambda d(h_t, h_0)] + \sqrt{\alpha}_t \cdot z
  \label{eq:iter_main}
  \end{equation}
where $\lambda$ is a hyperparameter that controls the trade-off between the likelihood term and the prior distribution. The trade-off between them helps to balance two competing objectives: steering the hidden state toward CoT reasoning mode while maintaining its proximity to the original language manifold.
The complete optimization procedure is presented in Algorithm~\ref{alg:cot_optimization}.

\subsection{Balancing Likelihood and Prior}
\label{sec:tradeoff}
Determining proper bounds for $\lambda$ is crucial, as inappropriate values may fail to reach the CoT mode or push hidden states off-manifold, resulting in incoherent text generation.
In this section, we establish theoretical bounds for $\lambda$ from two perspectives: ensuring gradient alignment with the likelihood objective, and maintaining appropriate step magnitude during update.

\textbf{Aligning Optimization Direction}
The gradient of the likelihood term $\nabla_h(\log f_\theta(h_t))$ directly points toward the CoT mode. 
For effective optimization, we expect the overall gradient of Eq.~\ref{eq:iter_main} to closely align with this CoT-inducing direction, which we achieve by maximizing their cosine similarity, ensuring the updates predominantly towards CoT mode.
Denote the gradient of $h$ in Eq.~\ref{eq:iter_main} is $\nabla^{**}$, and denote $\nabla((\log f_\theta(h_t)))$ is $\nabla^*$. 
% the cosine similarity is defined as
% \begin{equation}
%     \label{eq:def_lb}
%     Cos(\nabla^*,\nabla^{**}) = \frac{\nabla^* \cdot [\nabla^* - \nabla_h\left(d(h_t, h_0)\right)]}{\|\nabla^*\|\times\|\nabla^{**}\|},
%   \end{equation}
By maximizing the cosine similarity $Cos(\nabla^*,\nabla^{**})$, we can derive the upper bound of $\lambda$.

\begin{customlem}{1}
    \label{thm:ub}
       Assume that the cosine similarity is larger than $1-\epsilon_c$, where $\epsilon_c$ is a positive small number close to $0$. At any timestep $t$, $\lambda$ must be smaller than the upper bound as
       \begin{equation}
           \lambda \leqslant \frac{\epsilon_c\sum_{n}^{i=1}{\nabla^*_i}^2}{2\sum_{n}^{i=1}{h_t}_i \times \nabla^*_i}
       \end{equation}
       where $n$ is the length of $h_t$, and ${h_t}_i$ and $\nabla^*_i$ denotes the $i$-th item in $h_t$ and $\nabla^*$, respectively.
       \vskip -0.1in
\end{customlem}

The detailed proof of Lemma~\ref{thm:ub} is in the appendix. With the upper bound stands, $\lambda$ can be guaranteed to be a very small positive number. This helps keep each step of the gradient up on the correct path.

\textbf{Maximizing the Distance.}
Beyond proper directional alignment, we also need to calibrate step magnitude during optimization to prevent drifting outside from the  distribution.
Therefore, the distance between $d(h_{t+1}, h_0)$ is expected to get smaller at each iteration. At timestep $t$, we assume that $\hat{h}_{t+1}$ is a special case of $h_{t+1}$ where $\lambda = 0$ as
\begin{equation}
    \hat{h}_{t+1} = h_t+\alpha_t\nabla_h(\log f_\theta(h_t)) + \sqrt{\alpha_t}\cdot z.
\end{equation}
We can minimize $d(h_{ t+1}, h_0)$ by maximizing $d(\hat{h}_{t+1}, h_0) - d(h_{t+1}, h_0)$. Detailed proof is in the appendix.

\begin{customlem}{2}
 \label{thm:lb}
    Assume that $\|\hat{h}_{t+1}\| - \|h_{t+1}\|$ is larger than a positive number $\epsilon_d$. Ignoring the regularization term and denoting the cosine of angle $<h_{t+1}, \nabla_h(\|h_t\|)>$ as $C_t$, the lower bound of $\lambda$ can then be derived using the cosine theorem as
    \begin{equation}
      \lambda \geqslant \frac{\nabla^*+\|h_t\|\cdot C_t-\sqrt{[\nabla^*+\|h_t\|\cdot C_t]^2-\epsilon_t}}{\nabla_h(\|h_t\|)}
    \end{equation}
    See detailed proof in the appendix.
\end{customlem}

With Lemma~\ref{thm:ub} and Lemma~\ref{thm:lb} working together, the range of $\lambda$ is
\begin{align}
\frac{\nabla^*+\|h_t\|\cdot C_t-\sqrt{[\nabla^*+\|h_t\|\cdot C_t]^2-\epsilon_t}}{\nabla_h(\|h_t\|)} \leqslant \lambda \notag \\
\leqslant \frac{\epsilon_c\sum_{n}^{i=1}{\nabla^*_i}^2}{2\sum_{n}^{i=1}{h_t}_i \times \nabla^*_i}
\end{align}

With this constraint, the optimization maintains both update directional alignment and appropriate step size magnitude. 
Each optimization step ensures the hidden state moves efficiently toward the CoT reasoning mode while remaining within the proximity of the natural language distribution.

\begin{algorithm}[t]
    \caption{Hidden State Optimization for Chain-of-Thought Reasoning}
    \begin{algorithmic}[1]
    \Require A base $\text{LLM}$, trained classifier $f_\theta$, input tokens $X$, target layer $l$, trade-off $\lambda$, step size $\alpha_0$, threshold $\tau$, maximum iterations $T$, generation length $n$
    \Ensure Hidden stats are optimized to desired mode.
    
    % \State Initialize generated tokens $Y \gets []$
    \For{$i = 1$ to $n$}
        \State $h_0^{(l)} \gets \text{LLM.forward}(X, l)$ \Comment{Layer $l$ Forward }
        
        % \State $p_{\text{CoT}} \gets f_\theta(h^{(l)})$ \Comment{Classify if hidden state exhibits CoT}
        
        \If{$f_\theta(h^{(l)}) < 0.5$} \Comment{Optimize negative samples}
            % \State $h_0^{(l)} \gets h^{(l)}$ \Comment{Store original hidden state}
            \State $h_t^{(l)} \gets h_0^{(l)}$ \Comment{Initialize optimization}
            
            \For{$t = 1$ to $T$}
                \State $g \gets \nabla_{h}[\log f_\theta(h_t^{(l)}) - \lambda d(h_t^{(l)}, h_0^{(l)})]$ \Comment{Compute gradient}
                
                \State $\alpha_t \gets \alpha_0 \cdot \frac{|\tau - f_\theta(h_t^{(l)})|}{|f_\theta(h_0^{(l)})| + \epsilon}$ \Comment{Adaptive step length}
                
                % \State $z \sim \mathcal{N}(0, I)$ \Comment{Sample random noise}
                
                \State $h_{t+1}^{(l)} \gets h_t^{(l)} + \alpha_t g + \sqrt{\alpha_t} \cdot z$ \Comment{Update}
                
                \If{$f_\theta(h_{t+1}^{(l)}) > \tau$}
                    \State \textbf{break} \Comment{Stop when condition is satisfied}
                \EndIf
            \EndFor
            
            % \State $h^{(l)} \gets h_{t+1}^{(l)}$ \Comment{Replace with optimized hidden state}
        \EndIf
        
        \State $\text{LLM.update\_layer\_activation}(l, h_{t+1}^{(l)})$ 
        \State $x_i \gets \text{LLM.forward\_from\_layer}(l)$ \Comment{Update hidden state and continue forward pass}
        % \State $Y \gets Y \cup y_i$ \Comment{Append generated token}
        \State $X \gets X \cup x_i$ \Comment{Update input context}
    \EndFor
    
    \State \Return $X$ \Comment{Return generated text}
    \end{algorithmic}
    \label{alg:cot_optimization}
    \end{algorithm}

\section{Experiments}

\subsection{Setup}
\textbf{Datasets:} 
We evaluate our method on three reasoning task types: math, commonsense, and logical. 
For math, we select four well-known datasets: GSM8K~\cite{cobbe2021training}, MultiArith~\cite{roy2016solving}, SVAMP~\cite{patel2021nlp}, and MAWPS~\cite{koncel2016mawps}, which encompass a range of problem difficulties and structures.
We also choose two popular commonsense datasets: CommonsenseQA(C-QA)~\cite{talmor2018commonsenseqa} and StrategyQA(S-QA)~\cite{geva2021did} and two popular logical datasets: Objects Tracking~\cite{srivastava2022beyond} and Coin Flip~\cite{srivastava2022beyond}.
\textbf{LLMs:} We choose four solely pre-trained base LLMs: Mistral-7b~\cite{jiang2023mistral}, Gemma-7b~\cite{team2403gemma}, Qwen1.5-4b~\cite{bai2023qwen} and Phi-1.5~\cite{li2023textbooks}, running on NVIDIA 4090 GPUs.

% \usepackage{multirow}
\begin{table*}[ht]
  \centering
    \resizebox{0.8\textwidth}{!}{
      \begin{tabular}{cccccccccc}

        \hline
        \multirow{2}{*}{\centering LLM}                 & \multirow{2}{*}{\centering Method}  & \multicolumn{4}{c}{Math}                                                               & \multicolumn{4}{c}{Commonsense and Logic}                                                                                                         \\ \cline{3-10} 
       &     & GSM8K  & MultiArith   & SVAMP    & MAWPS     & C-QA     & S-QA      & Obj-T        & C-Flip             \\ \hline
\multirow{6}{*}{\centering Mistral-7b}   & Vanilla  & 11.32\%   & 15.55\%  & 52.66\%  & \multicolumn{1}{c|}{56.61\%}    & 56.06\%  & 50.59\%   & 33.08\%    & 48.75\%       \\
& C-DIM       & 13.42\%  & 17.86\%     & 54.76\%   & \multicolumn{1}{c|}{57.26\%}  & 55.47\%  & 50.31\%   & 32.84\%    & 49.65\%             \\
& C-PCA       & 14.52\%  & 17.65\%     & 55.24\%   & \multicolumn{1}{c|}{57.87\%}  & 56.32\%  & 51.85\%   & 33.83\%    & 48.49\%              \\
& C-LR         & 15.86\%  & 18.93\%     & 56.97\%   & \multicolumn{1}{c|}{58.97\%}  & 56.94\%  & 49.63\%   & 34.05\%    & 49.75\%             \\
& P-SVM       & 15.74\%  & 20.86\%     & 56.48\%   & \multicolumn{1}{c|}{\textbf{59.45\%}}   & 55.83\%  & 51.85\%   & 35.64\%   & 50.74\%      \\
& DA          & 16.83\%  & 19.35\%     & 55.14\%   & \multicolumn{1}{c|}{57.63\%}            & 54.21\%    & 50.04\%    & 32.45\%   & 49.35\%      \\
\multicolumn{1}{l}{} & Ours    & \textbf{18.24\%} & \textbf{23.33\%} & \textbf{57.33\%} & \multicolumn{1}{c|}{59.20\%} & \textbf{57.19\%}  & \textbf{52.04\%}  & \textbf{34.57\%}      & \textbf{50.87\%}             \\ \hline
\multirow{6}{*}{\centering Gemma-7b} & Vanilla  & 21.62\%    & 30.55\%   & 53.33\%    & \multicolumn{1}{c|}{64.44\%}   & 52.92\% & 42.83\%  & 31.68\%    & 45.54\%  \\
              & C-DIM   & 23.64\%    & 35.63\%   & 52.62\%    & \multicolumn{1}{c|}{65.58\%}   & 51.38\%    & 43.54\%  & 32.84\%     & 46.75\%                              \\
              & C-PCA   & 24.72\%    & 39.48\%   & 55.34\%   & \multicolumn{1}{c|}{65.74\%}     & 52.74\%   & 43.79\%  & 33.72\%    & 46.84\%                              \\
              & C-LR    & 26.76\%    & 39.74\%   & 55.25\%   & \multicolumn{1}{c|}{67.86\%}    & 53.41\%   & 42.86\%  & 32.31\%     & 47.71\%                              \\
            & P-SVM   & 26.32\%    & 40.61\%   & 57.83\%   & \multicolumn{1}{c|}{67.53\%}      & 53.56\%  & 43.61\%  & 32.86\%    & 46.53\%                              \\
              & DA      & 25.75\%    & 39.43\%   & 56.47\%   & \multicolumn{1}{c|}{66.49\%}      & 52.75\%  & 42.35\%  & 31.26\%  & 47.52\%                              \\
\multicolumn{1}{l}{} & Ours    & \textbf{28.79\%} & \textbf{45.11\%} & \textbf{59.67\%} & \multicolumn{1}{c|}{\textbf{69.92\%}} & \textbf{54.78\%}   & \textbf{44.42\%}    & \textbf{34.54\%}  & \textbf{48.81\%}   \\ \hline
\multirow{6}{*}{\centering Qwen1.5-4b}   & Vanilla  & 49.26\% & 89.00\%  & 54.33\%   & \multicolumn{1}{c|}{65.24\%}   & 71.67\%     & 53.41\%  & 28.54\%  & 52.78\%     \\
& C-DIM   & 50.63\%  & 88.74\%     & 55.14\%     & \multicolumn{1}{c|}{64.62\%}       & 72.74\%    & 54.52\%  & 30.39\%  & 53.41\%   \\
& C-PCA   & 49.74\%  & 89.78\%     & 55.45\%    & \multicolumn{1}{c|}{65.04\%}       & 71.09\%    & 55.75\%  & 31.42\%  & 52.63\%      \\
& C-LR    & 50.94\%  & 90.32\%     & 55.87\%     & \multicolumn{1}{c|}{66.13\%}      & 72.74\%     & 55.93\%  & 30.84\%  & 53.82\%                              \\
& P-SVM   & 51.24\%  & 90.15\% & 56.24\%   & \multicolumn{1}{c|}{67.45\%}    & 73.12\%    & 57.34\%  & 30.75\%  & 54.84\%                              \\
& DA      & 50.02\%  & \textbf{91.73\%} & 55.39\%   & \multicolumn{1}{c|}{66.56\%}    & 71.42\%    & 55.65\%  & 30.46\%  & 53.65\%                              \\
\multicolumn{1}{l}{} & Ours    & \textbf{51.33\%} & 91.66\%  & \textbf{57.33\%} & \multicolumn{1}{c|}{\textbf{68.62\%}} & \multicolumn{1}{l}{\textbf{73.48\%}} & \multicolumn{1}{l}{\textbf{59.52\%}} & \multicolumn{1}{l}{\textbf{32.45\%}} & \multicolumn{1}{l}{\textbf{55.63\%}} \\ \hline
\multirow{6}{*}{\centering Phi-1.5}   & Vanilla  & 5.69\%  & 7.78\%  & 11.67\%   & \multicolumn{1}{c|}{15.29\%}  & 27.93\%    & 37.53\%   & 31.60\%  & 49.34\%                              \\
& C-DIM   & 5.43\%   & 10.45\%     & 13.55\%    & \multicolumn{1}{c|}{15.84\%}    & 28.14\%    & 37.24\%   & 31.75\%   & 49.72\%                              \\
& C-PCA   & 5.84\%   & 11.53\%     & 12.64\%   & \multicolumn{1}{c|}{16.94\%}     & 29.31\%     & 37.85\%    & 32.48\%    & 50.04\%                              \\
& C-LR    & 6.47\%   & 13.75\%     & 13.93\%   & \multicolumn{1}{c|}{16.08\%}    & 28.75\%      & 38.14\%    & 31.04\%   & 50.75\%                              \\
& P-SVM   & 6.24\%   & 13.54\%     & 14.73\%   & \multicolumn{1}{c|}{16.75\%}    & 28.63\%      & 37.95\%  & 32.54\%    & 49.47\%                              \\
& DA      & 6.85\%   & 14.56\%     & 14.91\%   & \multicolumn{1}{c|}{16.31\%}    & 29.18\%      & 38.63\%  & 32.79\%    & 50.15\%                              \\
\multicolumn{1}{l}{} & Ours    & \textbf{6.74\%}  & \textbf{16.67\%} & \textbf{15.00\%} & \multicolumn{1}{c|}{\textbf{17.50\%}} & \textbf{29.84\%}   & \textbf{39.74\%}   & \textbf{34.33\%}  & \textbf{51.27\%}   \\ \hline
        \end{tabular}
    }
    \vspace{1mm}
  \caption{Problem-solving accuracy comparison between our method and baseline approaches across diverse reasoning tasks. Bold values indicate best performance for each model-dataset combination. Our approach consistently improves performance across nearly all models and datasets. Vanilla is defaultgreedy decoding performance.}
  \label{tab: main results}
  \vspace{-2mm}
  \end{table*}

\textbf{Classifier and Hidden states:}
To build the training dataset for our classifier, we first generate paired responses for each problem. 
We input problems from the GSM8K training set into Mistral-7b and sample multiple responses per problem. 
These responses are then classified by GPT-4 as either CoT (with explicit reasoning steps and correct solutions) or non-CoT (with direct answers or irrelevant content), resulting in a dataset of 2000 response pairs. 
This dataset is split into training and validation sets in an 8:2 ratio. This paired dataset is used to train classifiers for four base LLMs.

Next, to obtain the hidden states for training the classifiers, we concatenate the questions and responses and feed them into the LLM. 
At each layer of the LLM, we extract the hidden states corresponding to CoT and non-CoT responses. These hidden states are then used to train a classifier specific to each layer. 
Our classifier $f_\theta$ is designed as a two-layer MLP with ReLU activations and a sigmoid output. It is trained on these contrastive hidden states using cross-entropy loss to distinguish between CoT and non-CoT representations effectively. More classifier details are in the appendix.

Specifically, we consider three types of hidden representations in each transformer layer: the activations from the self-attention component(ATTN), the multi-layer perceptron(MLP) component, and the integrated layer output(INT-Layer). 
During optimization, we select the hidden state type with the best classification performance to ensure effective steering of the model's reasoning capabilities. Additionally, we choose the optimization layers from the top 50\% of layers based on their classification performance. 
The justification for this selection strategy is detailed in Section \ref{sec:hidden_states_and_layers}.

\subsection{Evaluation Metrics}
\textbf{Answer Accuracy}: 
The correctness of answers, which tends to be higher when CoT reasoning is present in the generated text.
\textbf{Fluency}: Weighted average of bi- and tri-gram entropies, calculated as $-\sum_k f(k) \log_2 f(k)$, where $f(\cdot)$ is the n-gram frequency and $k$ denotes each unique n-gram in the text~\cite{meng2022mass}.
\textbf{Perplexity}: 
Measure the model's ability to model the response, with higher values indicating weaker modeling capability and a shift in distribution during optimization.
\textbf{GPT4o Judge Score}: We use GPT-4o to evaluate the reasoning content in generated text, scoring from 0 (direct answers without reasoning) to 1 (detailed chain-of-thought content). The prompt is the appendix.

\subsection{Answer Accuracy Results}

\textbf{Config}: We set hyperparameters as follows: maximum iterations at 200, $\alpha_0$ at 0.1, $\lambda$ at 0.01, and $\tau$ at 0.9. Hyperparameter analysis is discussed in Section 4.4. For input formatting, we use a straightforward question-answer template: ``Question:[question]\verb|\n|Answer:" without additional prompting techniques.
For hidden states, we select integrated layer output for Mistral-7B and Gemma-7B, and self-attention activations for Qwen1.5-4b and Phi-1.5, based on their classification performance.

\textbf{Baselines:}
We compare action steering methods that modify hidden states with control vectors from various techniques:

- Difference in Mean(C-DiM) \cite{hojerimproving}: 
Constructs a control vector by calculating the difference between average representations of CoT and non-CoT examples.

- Principal Component Analysis(C-PCA) \cite{zou2023representation}: 
Creates a control vector by finding the principal component of differences between sampled CoT and non-CoT representations.

- Logistic Regression(C-LR) \cite{von2024language}: 
Trains a logistic regression model on CoT and non-CoT hidden states, using the weight vector as the control vector for reasoning direction.

We also compare activation editing methods:

- Boundary Projection in SVM(P-SVM) \cite{huang2025jam}: 
Trains an SVM to obtain a classification hyperplane, then projects negative sample representations onto this boundary surface.

- Directional Ablation(DA) \cite{arditi2024refusal}: 
Ablates the negative DiM control vector by projecting activations orthogonal to this direction, eliminating the model's tendency to refrain from reasoning.

All baseline methods intervene at the same layers as our method, and the strength of the control vectors is uniformly set to 1 for consistency in comparison.
For all experiments, the sampling strategy in text generation is set to greedy sampling.

\textbf{Analysis:} Table \ref{tab: main results} presents a comprehensive comparison of answer accuracy between our method and various baselines across multiple LLMs and tasks. 
The Vanilla baseline represents the performance of unmodified greedy sampling without any intervention on the model's hidden states.
Control vector-based methods achieve only marginal improvements in almost all tasks, likely due to their inability to precisely steer activations toward optimal regions and their lack of regularization during the steering process.

Our method, in contrast, demonstrates substantial improvements across nearly all LLMs and domains, with particularly impressive results on math reasoning benchmarks—achieving performance gains of up to 14.56\% on MultiArith using Gemma-7b and 8.89\% on MultiArith using Phi-1.5. 
These significant enhancements validate our approach's effectiveness in unlocking latent chain-of-thought reasoning capabilities inherent in base models.

While improvements on commonsense and logical reasoning tasks are more modest, they remain consistent, with an average gain of 1.8\% over other baselines. 
This performance differential suggests these tasks may demand more domain-specific knowledge beyond pure reasoning ability. 
Furthermore, all classifiers are trained on Mistral-7B data but generalize well to other models, showing strong method transferability.
% Also, classifier-based methods (C-LR, P-SVM, and Ours) outperform training-free vector-based approaches (C-DiM, C-PCA), as discriminative training better captures subtle differences between reasoning and non-reasoning states.

\begin{figure}[t]
  \centering
  \includegraphics[width=0.49\textwidth]{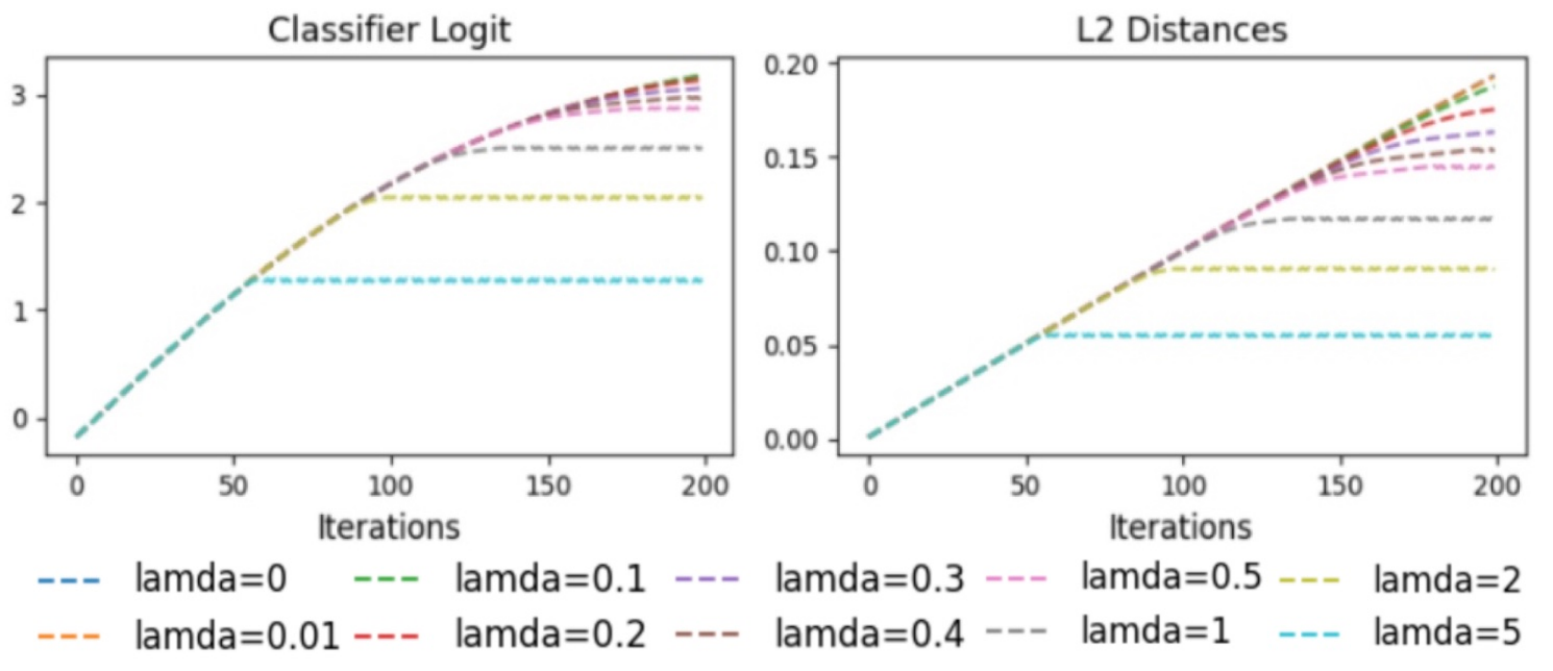}
  \caption{ Impact of different $\lambda$ values on the iterative optimization process, demonstrating how this hyperparameter influences the optimization convergence.}
  \vspace{-1mm}
  \label{fig:lambda_tradeoff}
\end{figure}

\begin{figure}[t]
  \centering
  \includegraphics[width=0.5\textwidth]{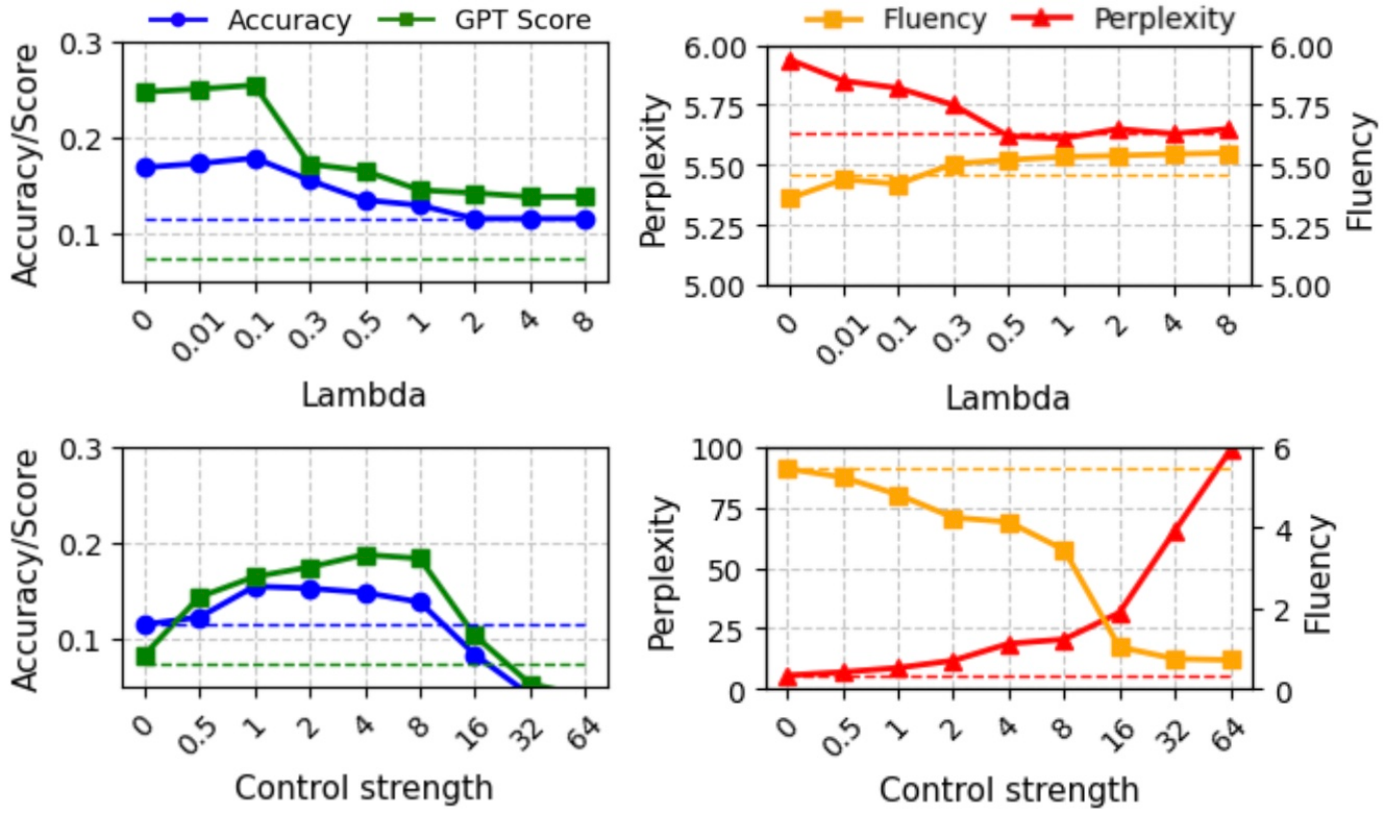}
  \caption{Upper: Impact of different $\lambda$ values on our method's performance, roughly showing the optimal range. Lower: Comparative analysis of activation steering with different strength. The dashed line in the figure represents the performance of the greedy decoding baseline.}
  \label{fig:ppl and score}
\end{figure}

\subsection{Optimization Config Analysis}
In this section, we analyze the sensitivity to hyperparameters like $\lambda$ and target threshold $\tau$. 
$\lambda$ balances likelihood and prior terms and $\tau$ sets the classifier output threshold.

\textbf{Trade-off between likelihood and prior terms:} 
Figure~\ref{fig:lambda_tradeoff} illustrates how varying $\lambda$ affects the optimization dynamics through three key metrics: logit value of classifier $f_\theta(\tilde{h})$, L2 distance $|h_t - h_0|^2$ across $\lambda$ values from 0 to 5.
% , and gradient norm $(|\nabla_h[\log f_\theta(\tilde{h}) - \lambda d(h_t, h_0)]|)$ 

When $\lambda$ is small (0-0.1), the likelihood term dominates the optimization, leading to a consistent increase in $f_\theta(h_t)$, with values of $f_\theta(h_t)$ being very close across different $\lambda$ in this range, while the L2 distance grows as $h_t$ diverges from the initial state $h_0$, similarly showing close values among different $\lambda$.
As $\lambda$ exceeds 0.1, the prior term becomes more influential, leading to quicker saturation of classifier output at lower values. 
Meanwhile, the L2 distance converges faster to smaller final values, accelerating the optimization process to earlier equilibrium.

Figure~\ref{fig:ppl and score} illustrates how our method's performance changes as the hyperparameter $\lambda$ varies. 
For $\lambda$ in the range of $0$-$0.01$, we observe an increase in performance alongside decreasing language quality, indicating an optimal operating region where the optimization process effectively balances the trade-off in Section~\ref{sec:tradeoff}. 
This sweet spot enables enhanced reasoning capabilities while maintaining high text quality.  
As $\lambda$ exceeds 0.01, language quality drops significantly, and reasoning performance also decline, indicating the prior term dominates, restricting exploration of effective reasoning paths. 
Nevertheless, across all $\lambda$ values, both metrics consistently surpass the vanilla baseline, showing enhanced reasoning capabilities regardless of hyperparameter settings.

For comparison, we plot the curves for C-LR, which uses an additive control vector with adjustable strength. Results show that C-LR's reasoning performance improves and language quality decreases for strength values between 0 and 2, yet it still underperforms compared to our method even at optimal strength.
Moreover, as strength increases, performance collapses below the greedy decoding baseline, and language quality deteriorates drastically, severely disrupting the model's output structure.
This highlights our method's superior stability and effectiveness across its parameter range.

\textbf{Evaluation of $\tau$:} 
In this evaluation, we assess the sensitivity of our method to $\tau$, the target threshold for classifier output, which is crucial for guiding the optimization process.

Table~\ref{tab:tau_analysis} show the results of varying $\tau$ from 0.5 to 0.99 to observe its impact, using Mistral-7b on the GSM8K dataset for validation. 
Here, $\tau=0.5$ represents the classification boundary. 
As $\tau$ increases from 0.5, reasoning performance, reflected in both accuracy and GPT score, steadily improves, reaching saturation around $\tau=0.9$, while language quality remains stable. 
When $\tau$ exceeds 0.9, the performance gains become marginal, indicating that further increases in the threshold beyond this point yield diminishing returns in reasoning improvement.

% \begin{table}[]
%     \caption{Analysis of different $\tau$ values on the iterative optimization process, using Mistral-7b on GSM8K}
%     \resizebox{0.5\textwidth}{!}{
%     \begin{tabular}{cccccc}
%     \hline
%     $\tau$  & Ave\_step  & Acc         & GPT-Score  & Fluency     & Perplexity \\ \hline
%     0.5     & 7          & 13.57\%     & 0.16        & 5.56       & 5.32          \\
%     0.6     & 11         & 14.56\%     & 0.18        & 5.46       & 5.42          \\
%     0.7     & 15         & 15.56\%     & 0.19        & 5.35       & 5.47          \\
%     0.8     & 17         & 16.56\%     & 0.23        & 5.33       & 5.53          \\
%     0.9     & 21         & 18.24\%     & 0.25        & 5.27       & 5.59          \\
%     0.92    & 21         & 18.24\%     & 0.25        & 5.25       & 5.61          \\
%     0.95    & 22         & 18.24\%     & 0.25        & 5.25       & 5.63          \\
%     0.99    & 22         & 18.45\%     & 0.25        & 5.24       & 5.63          \\ \hline
%     \end{tabular}
%     }
%     \end{table}

\begin{table}[]
    \resizebox{0.48\textwidth}{!}{
    \begin{tabular}{ccccc}
    \hline
    $\tau$    & Accuracy   & GPT-Score  & Fluency     & Perplexity \\ \hline
    0.50       & 13.57\%     & 0.16        & 5.56       & 5.32          \\
    0.60       & 14.56\%     & 0.18        & 5.46       & 5.42          \\
    0.70       & 15.56\%     & 0.19        & 5.35       & 5.47          \\
    0.80       & 16.56\%     & 0.23        & 5.33       & 5.53          \\
    0.90       & 18.24\%     & 0.25        & 5.27       & 5.59          \\
    0.92      & 18.26\%     & 0.26        & 5.25       & 5.61          \\
    0.95      & 18.35\%     & 0.26        & 5.25       & 5.63          \\
    0.99      & 18.45\%     & 0.27        & 5.24       & 5.65          \\ \hline
    \end{tabular}
    }
    \caption{Analysis of different $\tau$ values on the iterative optimization process, using Mistral-7b on GSM8K}
    \label{tab:tau_analysis}
    \end{table}

\subsection{Hidden States and Layer Analysis}
\label{sec:hidden_states_and_layers}
In this experiment, we examine the effect of varying hidden states and layers during representation optimization. 

Table~\ref{tab:hidden_states_and_layers} displays the answer accuracy for Mistral-7b and Gemma-7b across these variations, along with the classification F1 score for each hidden state. 
Layers are ranked by classification accuracy in descending order, and the top-k layers are chosen for optimization.

The results show a correlation between classification performance and optimization effectiveness. 
Both models achieve the greatest improvements with hidden states yielding the best classification results.
As the proportion of top-k layers rises from 10\% to 50\%, accuracy improves significantly, but beyond 50\%, gains become marginal, making additional layers less cost-effective due to increased computational demands. More comprehensive overhead analysis is provided in the appendix.

\begin{table}[]
    \centering
    \resizebox{0.48\textwidth}{!}{
        \begin{tabular}{ccccc}
            \hline
            LLM        & layers-Topk & MLP(81.23) & ATTN(83.45) & Int-Layer(91.62) \\ \hline
            \multirow{7}{*}{\rotatebox{90}{Mistral-7b}} 
                       & 10\%   & 12.64\%            & 13.42\%             & 15.86\%              \\
                       & 20\%   & 14.52\%            & 14.74\%             & 16.83\%              \\
                       & 30\%   & 15.24\%            & 15.51\%             & 17.86\%              \\
                       & 40\%   & 15.74\%            & 16.26\%             & 18.93\%              \\
                       & 50\%   & 15.74\%            & 17.86\%             & 18.24\%              \\
                       & 70\%   & 15.74\%            & 17.86\%             & 18.56\%              \\
                       & 90\%   & 16.83\%            & 18.93\%             & 18.78\%              \\ \hline
            LLM &  layers-Topk     & MLP(83.47) & ATTN(81.73) & Int-Layer(92.31) \\ \hline
            \multirow{7}{*}{\rotatebox{90}{Gemma-7b}}   
                       & 10\%   & 21.79\%            & 21.45\%             & 23.86\%              \\
                       & 20\%   & 22.37\%            & 22.26\%             & 24.73\%              \\
                       & 30\%   & 23.65\%            & 23.59\%             & 25.93\%              \\
                       & 40\%   & 24.34\%            & 24.07\%             & 27.46\%              \\
                       & 50\%   & 24.84\%            & 24.85\%             & 28.79\%              \\
                       & 70\%   & 24.98\%            & 25.14\%             & 29.14\%              \\
                       & 90\%   & 25.14\%            & 25.55\%             & 29.32\%              \\ \hline
            \end{tabular}
            }
    \caption{Optimization layers and hidden states analysis, using Mistral-7b and Gemma-7b in GSM8K. We select the top-k layers based on the classification performance of the MLP, ATTN, and Int-Layer hidden states. The average classification F1 scores for each hidden state type are annotated.}
    \label{tab:hidden_states_and_layers}
    \end{table}

% As an inference-time technique, the method introduces additional computation during the LLM's forward pass, primarily due to the iterative optimization of hidden states. We focus on how time consumption correlates with two key factors: the number of optimization steps (iterations, denoted as T) and the number of layers optimized. 

% \textbf{Experimental Setup:} We evaluate the computational overhead using the Mistral-7B model on the GSM8K dataset. 
% We vary the maximum number of iterations per hidden state optimization from 10 to 200. 
% We select subsets of layers for optimization, starting with the most discriminative layer and incrementally adding layers in groups of 4, ordered by descending classification accuracy as determined by F1 scores from validation data. This process continues up to a total of 32 layers (Mistral-7B's 32 layers).

% We analyze computational costs across methods using Mistral-7B on GSM8K with an NVIDIA 4090 GPU and Huggingface Transformers. 
% As Table \ref{tab:computation_times} shows, detecting negative hidden states and computing loss with backpropagation adds minimal overhead. 
% The main cost comes from hidden states optimization. 
% This inference-time computation produces longer responses and higher accuracy, indicating successful elicitation of CoT reasoning capabilities.

% \input{tables/table2.tex}

\section{Related Work}
\textbf{Representation Engineering in LLMs}
Representation engineering~\cite{zou2310representation} posits that LLM hidden states encode high-level concepts as causal factors, enabling output control through manipulation of these representations.
This approach has proven effective across domains: reducing hallucinations through truthfulness patterns~\cite{li2023inference, xiao2024enhancing, liu2024reducing}, rejecting harmful content via activation patterns~\cite{lee2024programming, wang2024locating}, modulating pesonality traits and political perspectives~\cite{zhu2024personality, kim2025linear}, and enhancing reasoning capabilities~\cite{hojer2025improving, wang2025thoughtprobe, hong2025reasoning, tang2025unlocking, hao2024training}.
Despite these advances, most are remain constrained by linear steering paradigms without sophisticated optimization frameworks.% for more principled representation manipulation.

\textbf{Reasoning Ability Enhancement in LLMs}
Methods to improve LLMs' reasoning abilities can be categorized into tuning-based and prompting-based approaches. Tuning-based methods rely on high-quality data and supervision.~\cite{zelikman2022star, zhang2024chain, hoffman2024training} bootstrap reasoning via iterative generation and filtering; Deepseek-R1~\cite{guo2025deepseek} employs outcome-based rewards to reinforce reasoning capabilities.Prompting-based methods utilize LLMs' few-shot learning~\cite{wei2022chain} and instruction-following abilities~\cite{yao2023tree, kojima2022large} to elicit reasoning patterns through carefully designed prompts. While effective, these approaches rely on external resources to enhance reasoning, rather than leveraging the intrinsic reasoning capabilities already encoded within LLMs' representations.

\section{Discussion and Limitations}
While effective, as an inference-time computation technique, our method can only unlock intrinsic reasoning abilities rather than inject new capabilities.
For models with limited intrinsic reasoning capacity established during pre-training, our approach cannot create non-existent capabilities. 
Additionally, our method's effectiveness depends significantly on the classification performance related to representations and layer selection; improper choices may lead to suboptimal results by failing to accurately capture the relevant reasoning states.
Looking ahead, we envision integrating our method with post-training techniques to further amplify the reasoning capabilities of LLMs.

\section{Conclusion}
We present a principled gradient-based activation optimization method to elicit CoT reasoning from base LLMs. 
Grounded in probabilistic generation theory, our approach derives an analytical objective from Bayesian principles that balances likelihood and prior terms—theoretically guaranteed to activate reasoning without causing distribution shift. Experiments across diverse reasoning benchmarks demonstrate our method consistently outperforms vector arithmetic approaches while maintaining textual coherence. This work offers a theoretically sound approach to enhance reasoning in foundation models without additional training.

\bibliography{aaai2026}


\section{Appendix}

\subsection{Theoretical Justification for the L2 Regularization Term}

In this appendix, we provide a detailed theoretical justification for the L2 regularization term used in the prior distribution \(p(h)\), establishing its connection to a Gaussian prior assumption. 
This derivation offers a statistical foundation for the optimization objective in Equation~(6) of the main paper, drawing from Bayesian perspectives on regularization in deep learning models~\cite{fortuin2022priors}.

\subsection{Gaussian Prior Assumption}
In Bayesian inference, the prior distribution \(p(h)\) encodes our beliefs about the hidden state \(h\) before observing the data. 
A common and computationally convenient choice is to model \(p(h)\) as a multivariate Gaussian distribution centered at the original hidden state \(h_0\), with covariance matrix \(\sigma^2 I\) (assuming isotropy for simplicity). 
The probability density function (PDF) of a \(d\)-dimensional Gaussian is given by:
\begin{equation}
p(h) = \frac{1}{(2\pi \sigma^2)^{d/2}} \exp\left( -\frac{1}{2\sigma^2} \|h - h_0\|^2 \right).
\label{eq:app_gaussian_pdf}
\end{equation}
This assumption implies that deviations from \(h_0\) are penalized quadratically, reflecting a belief in local smoothness around the initial state. 
Such Gaussian priors are widely used in variational inference and regularization techniques for neural networks, as they promote stability and prevent overfitting by favoring solutions close to the starting point~\cite{kingma2013auto, wilson2020bayesian}.

This Gaussian assumption is reasonable for hidden states in large language models (LLMs), as empirical studies show that activations in wide neural networks often approximate Gaussian processes in the infinite-width limit~\cite{lee2017deep}. 
For Transformer-based models, hidden states exhibit near-Gaussian marginal distributions in intermediate layers, especially after normalization~\cite{zhao2024beyond, skean2025layer}. 

\subsection{Derivation of the L2 Regularization Term}
To connect this to the optimization objective, we take the negative logarithm of the prior:
\begin{align}
-\log p(h) &= -\log \left[ \frac{1}{(2\pi \sigma^2)^{d/2}} \exp\left( -\frac{1}{2\sigma^2} \|h - h_0\|^2 \right) \right] \nonumber \\
&= \frac{d}{2} \log (2\pi \sigma^2) + \frac{1}{2\sigma^2} \|h - h_0\|^2.
\label{eq:app_neglog_prior}
\end{align}
Here, both \(\frac{d}{2} \log (2\pi \sigma^2)\) and \(\sigma^2\) (as a fixed hyperparameter) contribute constant factors that do not affect the optimization process. 
Thus, the relevant term is proportional to \(\|h - h_0\|^2\), which is equivalent to the L2 regularization in Equation~6 of the main paper, with the overall strength absorbed into a regularization parameter \(\lambda\).

In the MAP estimation framework (Equation~5 in the main paper), incorporating this prior into the posterior maximization objective effectively adds an L2 penalty:
\begin{align}
h^* &= \arg\max_h \left[ \log p(c|h) + \log p(h) \right] \notag \\ 
&\approx \arg\max_h \left[ \log p(c|h) - \lambda \|h - h_0\|^2 \right].
\label{eq:app_map_with_l2}
\end{align}

where \(\lambda\) encapsulates the scaling from \(\frac{1}{2\sigma^2}\) and other constants. Thus, the L2 term arises naturally from the Gaussian prior, constraining the optimization to a neighborhood around \(h_0\).

% \subsection{Justification in the Context of LLMs}

% However, we note that this is an approximation; real distributions may exhibit non-Gaussian traits like heavy tails, which could be addressed with alternative priors in future work~\cite{geifman2022heavy}.

\subsection{Proof of Lemma~1}
The target of Lemma~1 is to limit the cosine similarity to be not less than $1 - \epsilon_c$, where $\epsilon_c$ is a small and positive number. The target can be formulated as
\begin{equation}
    \label{eq:def_ub}
    \frac{\nabla^* \cdot \nabla^{**}}{\|\nabla^*\|\times\|\nabla^{**}\|} \geqslant 1-\epsilon_c.
\end{equation}

\begin{figure}[t]
    \centering
    \includegraphics[scale = 0.5]{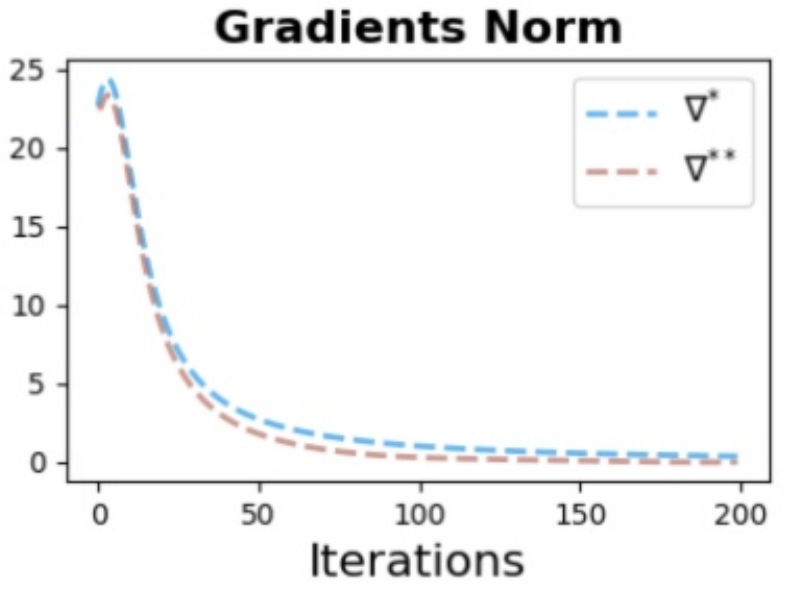}
    \caption{The gradient norm of $\nabla^*$ and $\nabla^{**}$ in the optimization process}
    \label{appfig:grad_norm}
    % \vspace{-2mm}
\end{figure}

In Eq.~\ref{eq:def_ub}, at each timestep $t$, $\|\nabla^{**}\|$ is observed to be changing in a very small range from $\|\nabla^*\|$, as shown in Fig~\ref{appfig:grad_norm}.
Thus, we can treat the two gradient norms as approximately equal for simplification.
$\nabla^{**}$ is composed of two independent functions related to $h$. By decomposing the gradient term, the formula can be reorganized as
\begin{equation}
    \label{eq:ub_reorg}
    \nabla^* \cdot [\nabla^* - \lambda\nabla_h\left(d(h_t, h_0)\right)] \geqslant \|\nabla^*\|^2(1-\epsilon_c).
\end{equation}
In this paper, $l_2$-distance is used to measure $d(h_t, h_0)$. Then at each item $i$, the gradient of $d(h_t, h_0)$ on ${h_t}_i$ is
% \begin{equation}
% \label{eq:nabla_d}
%     \frac{\left(d(h_t, h_0)\right)}{\partial {h_t}_i} = 2\times {h_t}_i
% \end{equation}

\begin{equation}
    \label{eq:nabla_d}
        \frac{\partial d(h_t, h_0)}{\partial {h_t}_i} = 2\times {h_t}_i
    \end{equation}
After expanding the matrix product and $l_2$-distances in Eq.~\ref{eq:ub_reorg}, with Eq.~\ref{eq:nabla_d}, we can calculate the inequality with the sum of each item. Eq.~\ref{eq:ub_reorg} can then be reorganized as

\begin{equation}
    \label{eq:ub_open}
    \sum_{n}^{i=1}{\nabla^*_i}^2 - \sum_{n}^{i=1}\left(\nabla^*_i \times 2\lambda{h_t}_i\right) \geqslant \sqrt{\sum_{n}^{i=1}{\nabla^*_i}^2}^2(1-\epsilon_c).
\end{equation}

Since lambda is invariant at any $i$, rearranging the formula makes it easy to put lambda on one side of the inequality as

\begin{equation}
    \label{eq:ub_def}
    \lambda \leqslant \frac{\epsilon_c\sum_{n}^{i=1}{\nabla^*_i}^2}{2\sum_{n}^{i=1}{h_t}_i \times \nabla^*_i}.
\end{equation}

This consists with the upper bound in Lemma~1. It can be seen that, with the upper bound of $\lambda$ being tightened via a smaller $\epsilon_c$, the cosine similarity is also bounded to be higher. When the cosine similarity is equal to 1, the upper level of the $\lambda$ must be equal to $0$, where the equal sign of Eq.~\ref{eq:ub_def} holds.

\subsection{Proof of Lemma~2}
Assume that the reference coordinate system is at $h_0=\mathbf{0}$, $d(\hat{h}_{t+1}, h_0) - d(h_{t+1}, h_0)$ can be represented by $\|\hat{h}_{t+1}\|| - \|h_{t+1}\|$. The target of Lemma~2 is to satisfy the constraint, 
\begin{equation}
\label{eq:lb_1}
    \|\hat{h}_{t+1}\| - \|h_{t+1}\| \leqslant \epsilon_d.    
\end{equation}
For simplification, we ignore $\alpha_t$. Since $h_{t+1} = \hat{h}_{t+1} + \lambda\nabla_h(\|h_t\|) $, by applying the cosine theorem, Eq.~\ref{eq:lb_1} can be expanded as

\begin{equation}
\label{eq:lb_2}
[\lambda\nabla_h(\|h_t\|)]^2-2\lambda\nabla_h(\|h_t\|)(\|h_t\|+\nabla^*)^2C_t \leqslant \epsilon_d
\end{equation}

It can be easily discovered that Eq.~\ref{eq:lb_2} is a quadratic function on $\lambda$. The solution of this function is the boundary on $\lambda$, which is

\begin{align}
\label{eq:lb_3}
\frac{\nabla^*+\|h_t\|\cdot C_t-\sqrt{[\nabla^*+\|h_t\|\cdot C_t]^2-\epsilon_t}}{\nabla_h(\|h_t\|)} \leqslant \lambda \leqslant \notag \\ 
\frac{\nabla^*+\|h_t\|\cdot C_t+\sqrt{[\nabla^*+\|h_t\|\cdot C_t]^2-\epsilon_t}}{\nabla_h(\|h_t\|)}
\end{align}

Since in Lemma~1, $\lambda$ has been defined in an interval for which an upper bound exists. Therefore, the final domain of definition of $\lambda$ should be the intersection of these two intervals. Since both $\epsilon_t$ and $\epsilon_c$ are defined to be a very small number, the upper bound in Lemma~1 is clearly smaller than the upper bound in Eq.~\ref{eq:lb_3}. So as Lemma~1 holds, $\lambda$ should only be bounded to a lower bound for Lemma~2. Finally, the lower bound of $\lambda$ is
\begin{equation}
    \lambda \geqslant \frac{\nabla^*+\|h_t\|\cdot C_t-\sqrt{[\nabla^*+\|h_t\|\cdot C_t]^2-\epsilon_t}}{\nabla_h(\|h_t\|)}
\end{equation}
As $\epsilon_t$ gets smaller, the lower bound of $\lambda$ gets lower. Until $\epsilon_t$ is $0$, the lower bound of $\lambda$ is $0$. At this point $\nabla_h(\|h_t\|)$ is $0$, consistent with the extreme case of the upper bound in Lemma~1.

\subsection{Classifier Training Details}

\subsubsection{Classifier Training Data Construction Details}
\label{app: classifier training data}
To train our classifier, we needed to collect distinct CoT and non-CoT responses for each question. Following~\cite{wang2024chain}, we implemented Top-K-Start Greedy Decoding, which explores the top-k alternative tokens at the first decoding step followed by greedy decoding thereafter, generating 10 diverse responses per question.
To ensure response quality, we applied a filtering process. 
We extracted only the solution-relevant content through post-processing, ensuring each response contained purely reasoning and answer components without redundant question restatements.
After this filtering, we employed GPT-4o as an expert judge to classify each filtered response as either CoT or non-CoT, creating a high-quality labeled dataset for our classifier training.
The prompt is shown in Fig.~\ref{appfig:classifier_date_prompt}:

\begin{figure*}[t]
    \centering
    \includegraphics[scale = 0.65]{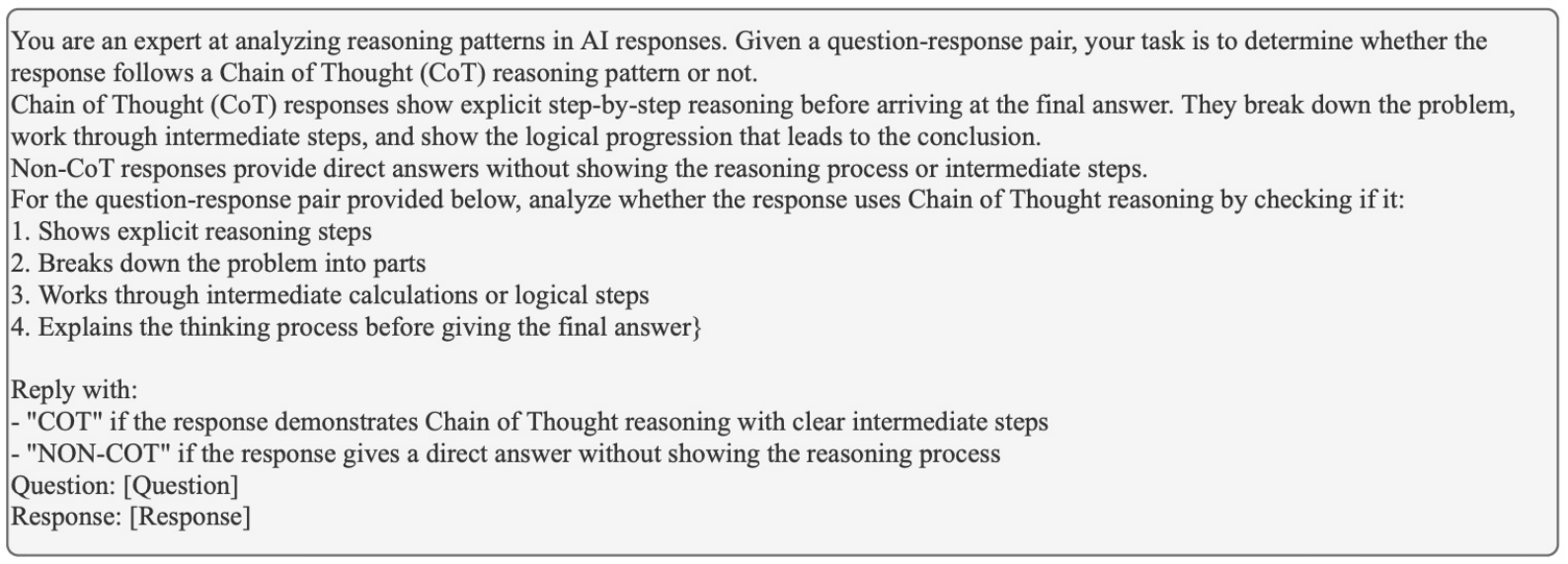}
    \caption{The prompt for GPT-4o to classify the response as CoT or non-CoT}
    \label{appfig:classifier_date_prompt}
\end{figure*}

We constructed domain-specific datasets for our classifier training:
For the mathematics domain, we collected questions from the GSM8K training set and sampled corresponding CoT and non-CoT responses. This resulted in a comprehensive dataset comprising 2,000 samples, with each sample representing a question-response pair.
For the commonsense-logical domain, we aggregated questions from multiple sources: the training sets of CommonsenseQA and StrategyQA, along with a separate collection from CoinFlip and ObjectTrack. 
Using these questions, we generated and labeled responses, ultimately creating a dataset of 2,000 samples.
Each dataset was carefully balanced to ensure robust classifier training across different reasoning contexts and patterns.

\subsubsection{Classifier Training Settings}
We extracted token-wise hidden representations from the LLM network and trained separate classifiers for each layer and representation type within the transformer architecture.
Specifically, we analyzed three types of hidden representations in each transformer layer: the activations from the self-attention component, the multi-layer perceptron (MLP) component, and the integrated layer output.
To address the length imbalance between positive and negative samples (where CoT responses tend to be longer due to their explicit reasoning steps, while non-CoT responses are typically more concise), we implemented a strategic sampling approach: extracting hidden representations every five tokens for CoT responses and every token for non-CoT responses. 
This created a more balanced training dataset while preserving representational diversity.
Our classifier architecture consists of a two-layer MLP with a ReLU activation function and a sigmoid output layer. 
We trained the model using cross-entropy loss, treating CoT data as positive samples and non-CoT data as negative samples. Training parameters included 100 epochs with a learning rate of 0.001, using stochastic gradient descent (SGD) as the optimizer.

\subsubsection{Classifier Performance}
\begin{figure*}[b]
    \centering
    \includegraphics[scale = 0.7]{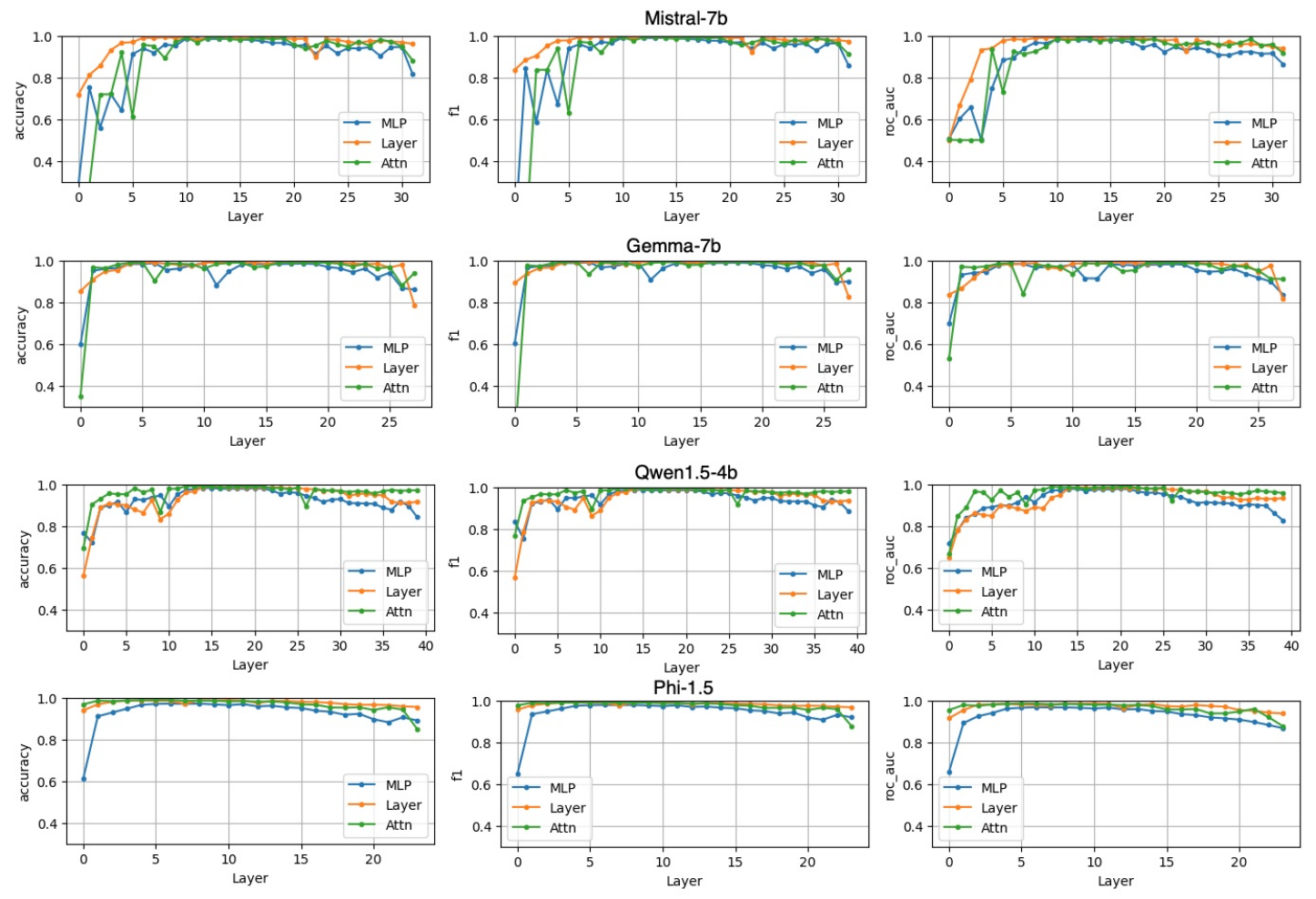}
    \caption{The classifier performance on math domain dataset}
    \label{appfig:classifier_performance on math}
\end{figure*}

\begin{figure*}[h]
    \centering
    \includegraphics[scale = 0.7]{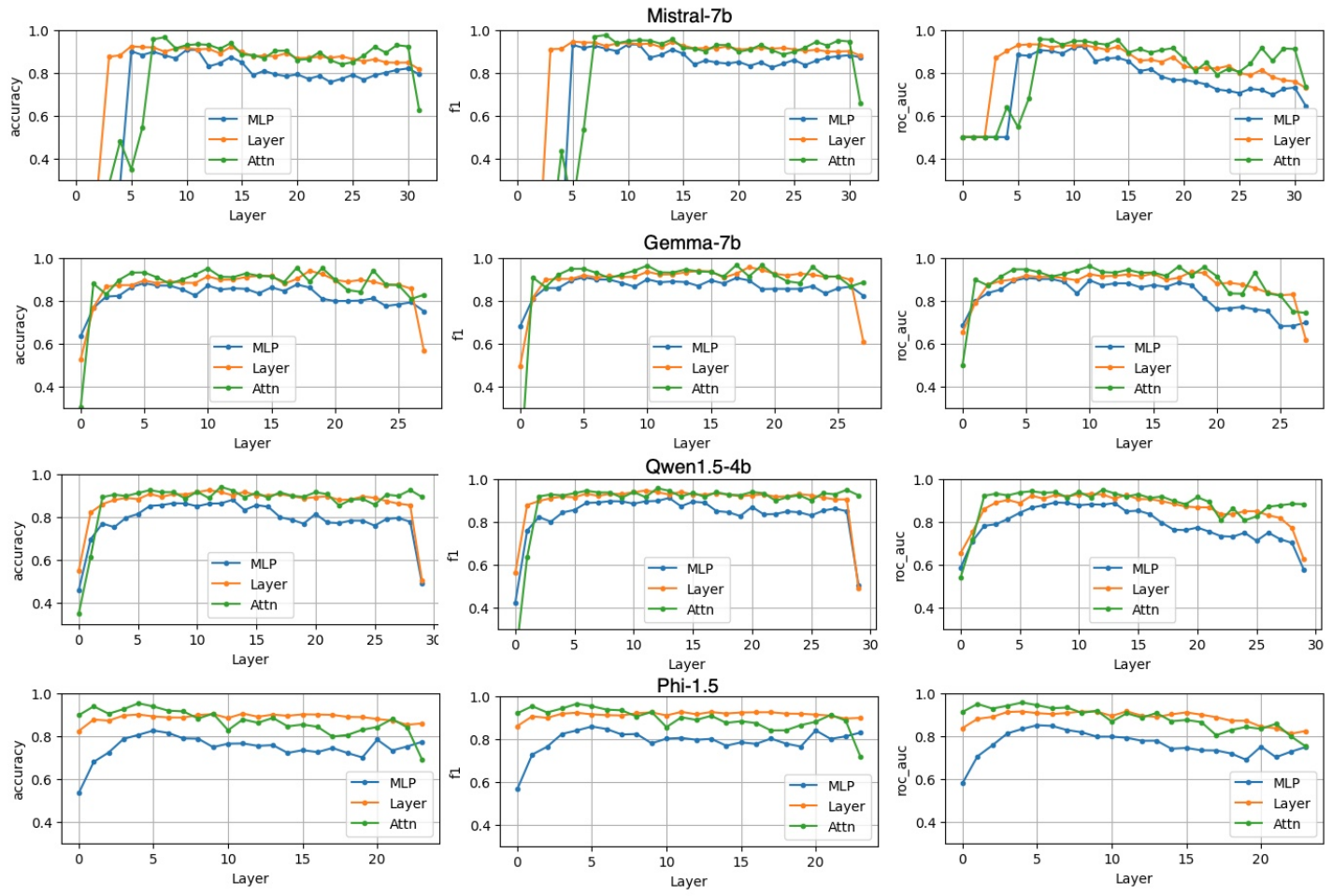}
    \caption{The classifier performance on commonsense-logical domain dataset}
    \label{appfig:classifier_performance on commonsense-logical}
\end{figure*}

We evaluated the classifier performance using multiple metrics: accuracy, ROC-AUC, and F1 score, providing a comprehensive assessment of its discriminative capabilities.
Figure~\ref{appfig:classifier_performance on math} and Figure~\ref{appfig:classifier_performance on commonsense-logical} illustrate the classifier's performance across math and commonsense-logical datasets, respectively. Our analysis revealed significant patterns in the classification efficacy across different network architectures.

Across all model variants, we observed a consistent pattern where middle layers demonstrated superior classification performance compared to both shallow and deep layers. This suggests that mid-level representations in transformer architectures capture reasoning patterns more effectively, while early layers may be too focused on syntactic features and deeper layers on task-specific outputs.

Interestingly, we found architecture-specific variations in which representation types yielded optimal performance. For Mistral-7B and Gemma-7B, the integrated layer representations provided the strongest discriminative signals. In contrast, Qwen1.5-4B and Phi1.5 showed better performance when classifiers were trained on attention activation patterns. These differences highlight how reasoning capabilities manifest differently across model architectures and suggest that optimal steering strategies may need to be tailored to specific model families.

\subsubsection{Representation Selection for Optimization}
We strategically selected the optimal representation type and layers for our optimization process based on the classifier performance analysis.
For Mistral-7B and Gemma-7B, we selected the integrated layer representations, as these demonstrated superior discriminative power between CoT and non-CoT patterns in these architectures. 
Conversely, for Qwen1.5-4B and Phi1.5, we targeted the attention activation representations, which yielded the strongest classification signals for these models.
For optimization layer, we select the top 50\% classification performance layer.

Regarding layer selection, we focused our optimization efforts on the top-performing 50\% of layers as measured by classification accuracy. This approach concentrates computational resources on the network components most likely to influence reasoning capabilities, while avoiding manipulation of layers showing minimal discriminative potential between reasoning patterns.

\subsection{Representation Optimization Details}
Our optimization process operates across two distinct stages of language model operation:

\textbf{Prefilling stage}: During prompt processing, we implement batch optimization. We first identify all negative hidden states (those classified as non-CoT) and group them as a batch. We then compute the optimization loss for this batch collectively and perform gradient descent to optimize these representations simultaneously. 
The optimization process terminates when either 95\% of the hidden states are classified as positive (indicating successful steering toward CoT representations) or when we reach the maximum allowed optimization steps.

\textbf{Decoding stage}: During the response generation process, we evaluate intermediate representations using our classifier before generating each token. We only apply optimization when these representations are classified as negative, indicating non-CoT patterns. This selective approach enables us to intervene only when the model deviates from CoT reasoning patterns, thereby making the optimization process more efficient while ensuring that the generated response maintains consistent reasoning characteristics throughout the generation process.
\subsection{Evaluation Details}

\subsubsection{Answer Extraction from Responses}
To systematically extract precise answers from generated responses, we implemented a prompt-based extraction approach tailored to each domain.
For mathematical problems, we appended the prompt \textit{"Therefore, the answer (arabic numerals) is"} to efficiently extract numerical answers from potentially lengthy reasoning chains.

For commonsense-logical problems, we employed domain-appropriate prompts based on the question format. For binary judgments, we used \textit{"Therefore, the answer (Yes or No) is"}, while for multiple-choice questions, we applied \textit{"Therefore, among A through E, the answer is"}. 
This approach ensured consistent and accurate answer extraction across different reasoning contexts and question types.

\subsubsection{Prompt GPT-4o to judge the quality of a response}
\begin{figure*}[h]
    \centering
    \includegraphics[scale = 0.65]{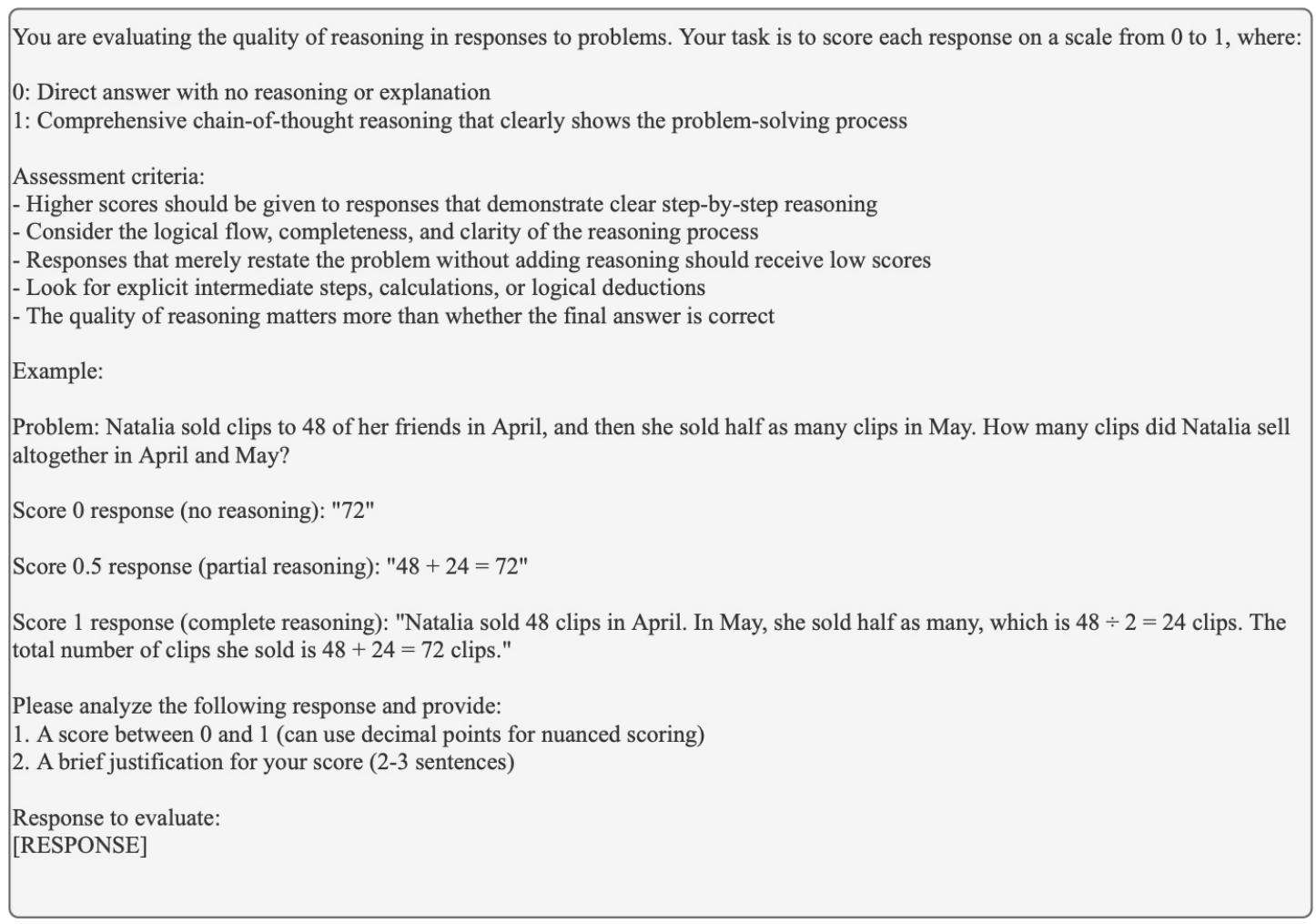}
    \caption{The prompt for GPT-4o to judge the quality of the response}
    \label{appfig:judge_response_prompt}
\end{figure*}

We prompt GPT-4o to judge the quality of the response by the following prompt shown in Fig.~\ref{appfig:judge_response_prompt}. 
The prompt instructed GPT-4o to assess each response on a normalized scale from 0 to 1, where higher scores indicate superior reasoning quality. 
We add an example to illustrate different response qualities and their corresponding scores, allowing for more reliable and consistent quality assessment across all evaluated samples.

\subsubsection{Computational Overhead}

We conducted a detailed analysis of computational costs across various methods using the Mistral-7B model on the GSM8K dataset, with experiments performed on an NVIDIA 4090 GPU utilizing the Huggingface Transformers library. 
As illustrated in Table~\ref{tab:computation_times}, the processes of detecting negative hidden states and computing loss with backpropagation introduce only minimal overhead, contributing negligibly to the overall computational burden. 
The predominant expense in terms of computation arises from the hidden states optimization phase, which is the core of our approach. 
To provide a clearer perspective on the efficiency of our method, we also measured the generation time per token, which underscores the computational impact of the optimization process on the inference pipeline. 
Despite the additional computational demands, this inference-time computation results in notably longer responses paired with improved accuracy, serving as a strong indicator of the successful elicitation of chain-of-thought (CoT) reasoning capabilities within the model.

\begin{table*}[]
    \centering
    \resizebox{0.8\textwidth}{!}{
        \begin{tabular}{cccc}
            \hline
            Method          & Process                       & Time(seconds) & Output Length \\ \hline
            Greedy decoding & Normal Generation             & 0.04          & 16             \\ \hline
            C-LR            & Adding a control vector       & 0.11          & 32             \\ \hline
            \multirow{2}{*}{\centering P-SVM} & Detect negative hidden states & 0.002         & \multirow{2}{*}{\centering 43}     \\
                            & Projection                    & 0.14          &                \\ \hline
            \multirow{3}{*}{\centering Ours} & Detect negative hidden states & 0.002         & \multirow{3}{*}{\centering 68}     \\
                            & Computing loss and backward   & 0.005         &                \\
                            & Hidden states optimization    & 0.33          &                \\ \hline
        \end{tabular}
    }
    \vspace{1mm}
  \caption{Average computation time per stage and response length for different approaches. While our method requires additional optimization time and produces longer responses, this reflects the successful generation of chain-of-thought reasoning.}
  \label{tab:computation_times}
  \vspace{-4mm}
  \end{table*}

\subsection{Case Study}
\begin{figure*}[h]
    \centering
    \includegraphics[scale = 0.65]{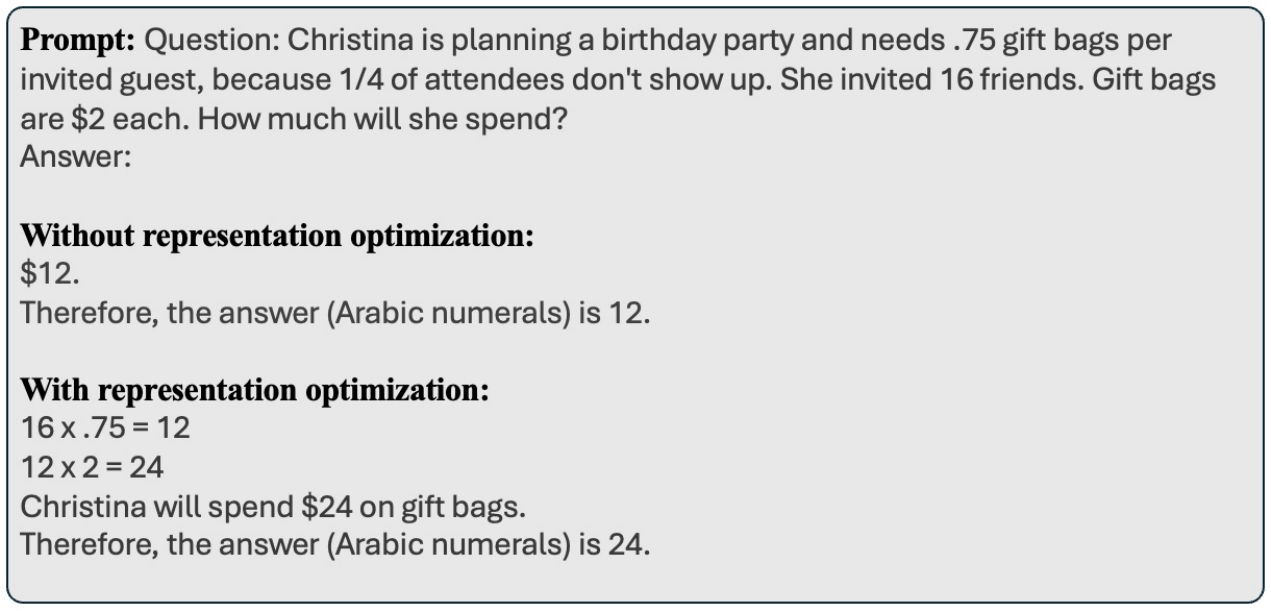}
    \caption{Case study 1}
    \label{appfig:case1}
\end{figure*}

\begin{figure*}[h]
    \centering
    \includegraphics[scale = 0.65]{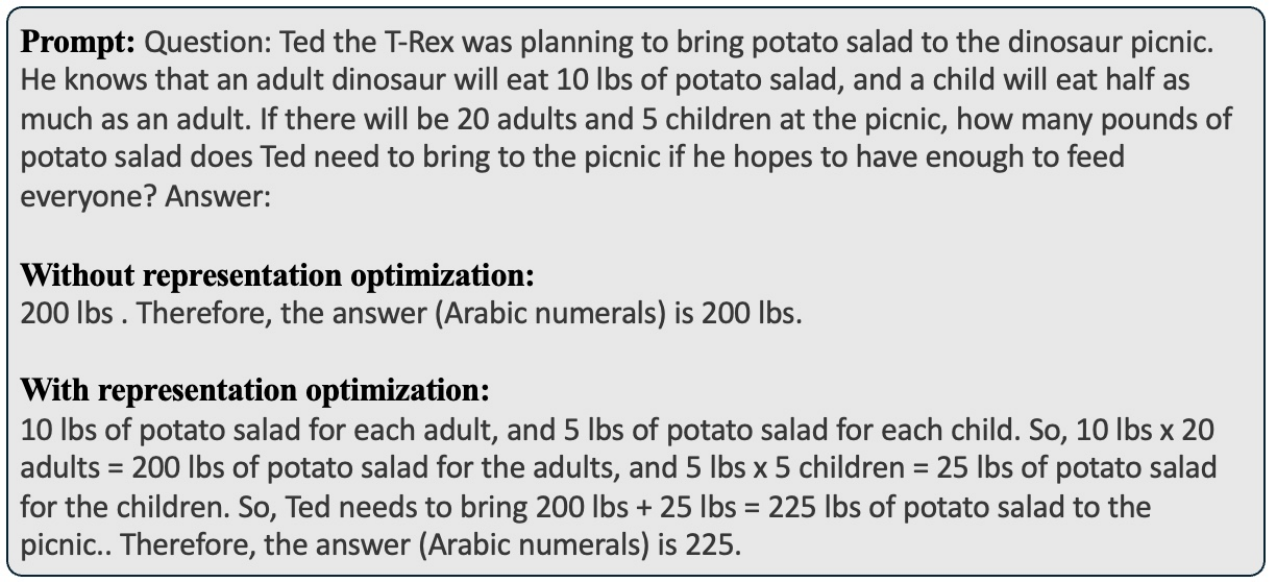}
    \caption{Case study 2}
    \label{appfig:case2}
\end{figure*}

\begin{figure*}[h]
    \centering
    \includegraphics[scale = 0.65]{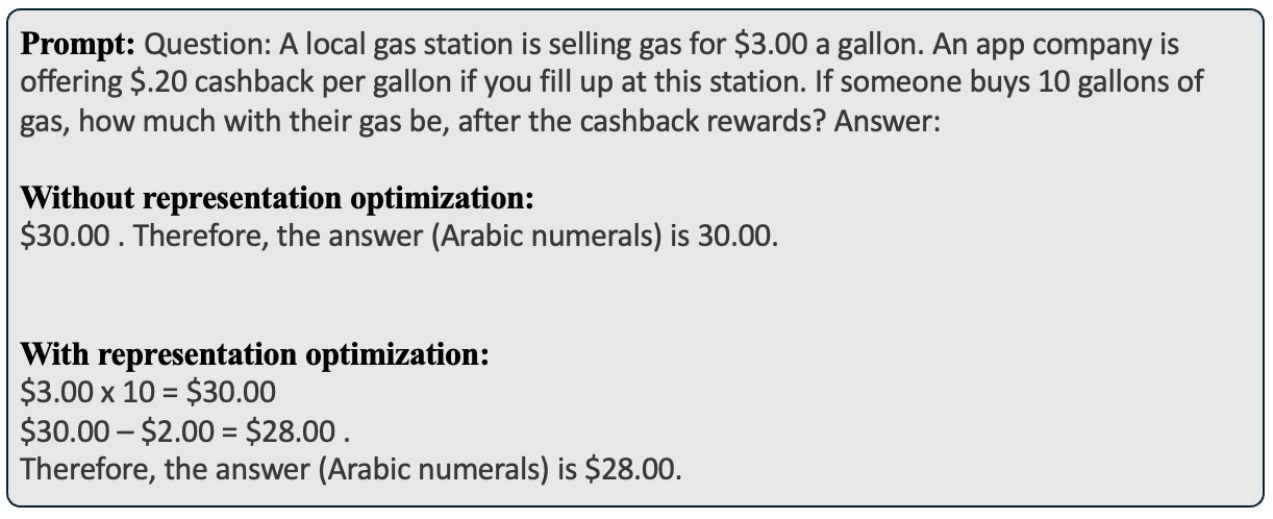}
    \caption{Case study 3}
    \label{appfig:case3}
\end{figure*}

\begin{figure*}[h]
    \centering
    \includegraphics[scale = 0.65]{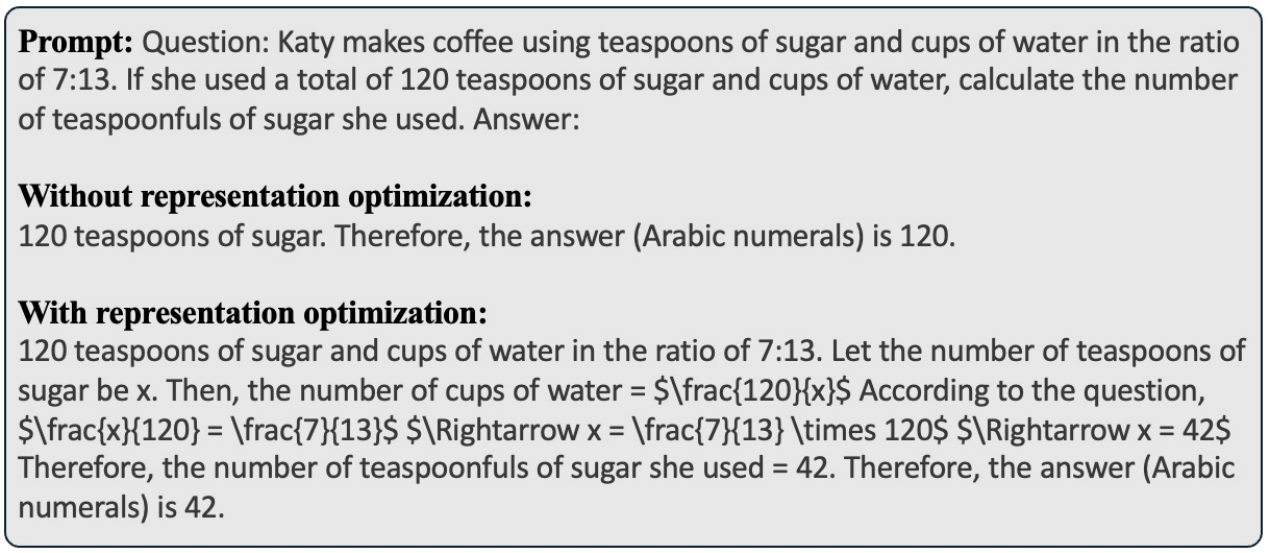}
    \caption{Case study 4}
    \label{appfig:case4}
\end{figure*}

\begin{figure*}[h]
    \centering
    \includegraphics[scale = 0.65]{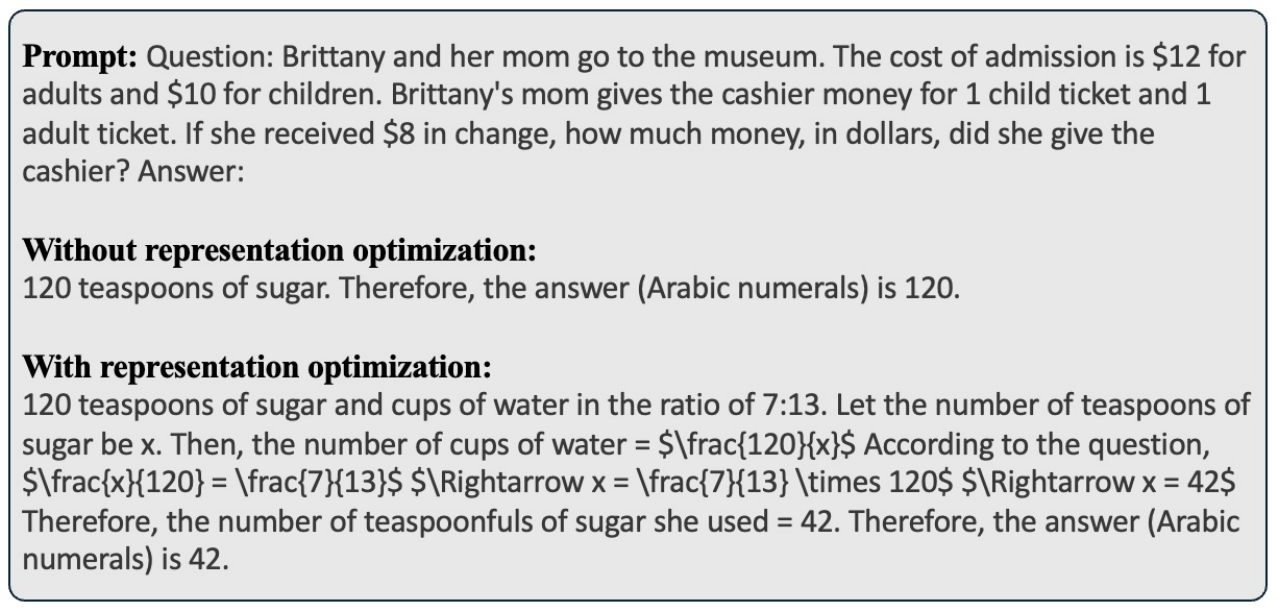}
    \caption{Case study 5}
    \label{appfig:case5}
    % \vspace{-2mm}
\end{figure*}

In Figures \ref{appfig:case1}, \ref{appfig:case2}, \ref{appfig:case3}, \ref{appfig:case4}, and \ref{appfig:case5}, we present detailed case studies from the GSM8K dataset using Mistral-7B to demonstrate the effectiveness of our representation optimization approach.

The results clearly illustrate the qualitative difference in reasoning patterns: without representation optimization, the model typically generates non-CoT responses that provide direct wrong answers. 
In contrast, when our representation optimization method is applied, the model produces comprehensive correct Chain-of-Thought responses that articulate a step-by-step reasoning process, revealing the model's internal problem-solving approach and enhancing solution transparency and accuracy.

These examples highlight how our method effectively unlocks the latent reasoning capabilities inherent in the base model without requiring additional training or external data.

\end{document}